\documentclass[10pt,twocolumn,letterpaper]{article}

\usepackage{cvpr}              
\usepackage{algorithm}
\usepackage{multirow}
\usepackage{float}
\usepackage{array} 

\definecolor{cvprblue}{rgb}{0.21,0.49,0.74}
\usepackage[pagebackref,breaklinks,colorlinks,allcolors=cvprblue]{hyperref}

\usepackage{algorithm}
\usepackage{algpseudocode}
\usepackage{amsmath}
\usepackage{adjustbox}
\usepackage{multirow}

\def\paperID{17603} 
\def\confName{CVPR}
\def\confYear{2025}

\usepackage{listings}
\usepackage{color}
\usepackage{float}
\usepackage{wrapfig}

\usepackage[accsupp]{axessibility}

\title{Rate-In: Information-Driven Adaptive Dropout Rates for Improved Inference-Time Uncertainty Estimation  \thanks{© 2025 IEEE. Personal use of this material is permitted. Permission from IEEE must be obtained for all other uses, in any current or future media, including reprinting/republishing this material for advertising or promotional purposes, creating new collective works, for resale or redistribution to servers or lists, or reuse of any copyrighted component of this work in other works. To appear in the Proceedings of the IEEE/CVF Conference on Computer Vision and Pattern Recognition (CVPR), 2025.}}

\author{
  Tal Zeevi$^{1}$ \quad
  Ravid Shwartz-Ziv$^{2}$ \quad
  Yann LeCun$^{2,3}$ \quad
  Lawrence H. Staib$^{1,\dag}$ \quad
  John A. Onofrey$^{1,}$\thanks{These authors contributed equally to this work.}
  \\[2pt
  ]
  $^{1}$Yale University \quad
  $^{2}$New York University \quad
  $^{3}$Meta FAIR
}

\begin{document}
\maketitle

\begin{abstract} Accurate uncertainty estimation is crucial for deploying neural networks in risk-sensitive applications such as medical diagnosis. Monte Carlo Dropout is a widely used technique for approximating predictive uncertainty by performing stochastic forward passes with dropout during inference. However, using static dropout rates across all layers and inputs can lead to suboptimal uncertainty estimates, as it fails to adapt to the varying characteristics of individual inputs and network layers. Existing approaches optimize dropout rates during training using labeled data, resulting in fixed inference-time parameters that cannot adjust to new data distributions, compromising uncertainty estimates in Monte Carlo simulations.

In this paper, we propose \textbf{Rate-In}, an algorithm that dynamically adjusts dropout rates during inference by quantifying the information loss induced by dropout in each layer's feature maps. By treating dropout as controlled noise injection and leveraging information-theoretic principles, Rate-In adapts dropout rates per layer and per input instance without requiring ground truth labels. By quantifying the functional information loss in feature maps, we adaptively tune dropout rates to maintain perceptual quality across diverse medical imaging tasks and architectural configurations. Our extensive empirical study on synthetic data and real-world medical imaging tasks demonstrates that Rate-In improves calibration and sharpens uncertainty estimates compared to fixed or heuristic dropout rates without compromising predictive performance. Rate-In offers a practical, unsupervised, inference-time approach to optimizing dropout for more reliable predictive uncertainty estimation in critical applications. 
\end{abstract}    
\section{Introduction}
\label{sec:intro}

\begin{figure*}[t]
    \includegraphics[width=\linewidth]{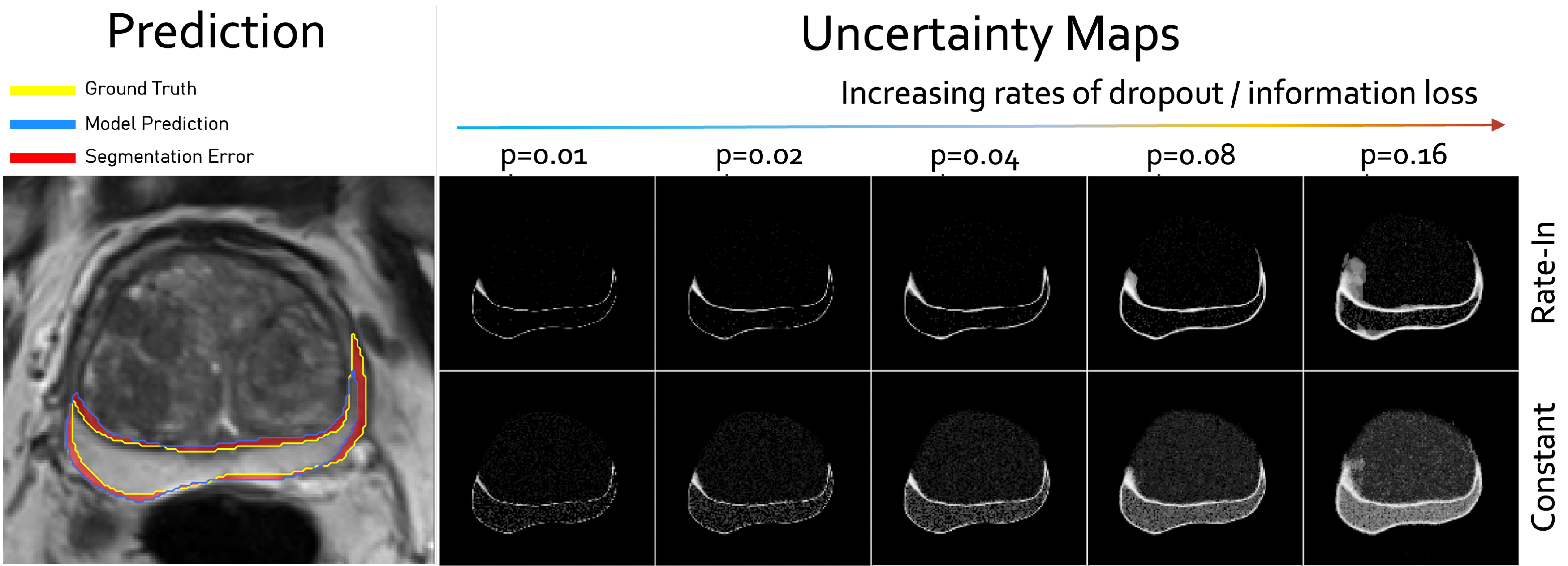}
    \caption{\textbf{Rate-In maintains anatomically meaningful uncertainty estimates at high dropout rates, while standard methods degrade into noise.} 
    Left: Prostate MRI segmentation showing ground truth (yellow), model predictions (blue), and errors (red). 
    Right: Uncertainty maps (brighter regions indicate higher uncertainty) at increasing dropout rates (1\%-16\%). 
    While constant dropout produces diffuse uncertainty that obscures anatomical details, Rate-In precisely highlights challenging regions, particularly along tissue boundaries, preserving the clinical utility of  uncertainty estimates.}

    \label{fig:dropout_viz}
\end{figure*}

Deep neural networks have achieved remarkable success in a wide range of applications, from image recognition to natural language processing. However, their deployment in risk-sensitive domains such as medical diagnosis, autonomous driving, and financial forecasting requires not only accurate predictions but also reliable estimates of predictive uncertainty~\cite{begoli2019need}. Accurate uncertainty estimation enables models to express confidence in their predictions, allowing practitioners to make informed decisions and to identify cases where the model may be less reliable.

Monte Carlo (MC) Dropout~\cite{gal2016dropout} is a practical technique for approximating predictive uncertainty in neural networks. By running stochastic forward passes with dropout activated during inference, MC Dropout samples from an approximate posterior distribution over the network's weights, enabling  estimation of the mean prediction and the associated uncertainty. This method offers a balance between computational efficiency and uncertainty estimation quality, making it suitable for large-scale applications.

A common limitation of MC Dropout is the use of fixed dropout rates during inference, typically the same rates used during training. These static dropout rates are applied uniformly across all layers and inputs, regardless of the varying characteristics of individual inputs and the differing sensitivities of network layers. This rigidity can lead to suboptimal uncertainty estimates~\cite{osband2016risk}: fixed dropout rates may either inject excessive noise, degrading the model's predictive performance, or fail to introduce sufficient variability to capture the true predictive uncertainty.

As shown in Figure~\ref{fig:dropout_viz}, how uncertainty is estimated can dramatically affect our ability to identify challenging regions in tasks such as medical image segmentation. Traditional uncertainty estimation methods tend to produce overly diffuse uncertainty patterns, failing to separate genuinely difficult decisions from simpler ones. This limitation becomes particularly critical in medical applications, where highlighting truly uncertain regions can guide clinical attention to areas requiring additional scrutiny.

Existing approaches have explored optimizing dropout rates during training~\cite{ba2013adaptive,gal2017concrete,kingma2015variational}, often relying on labeled data to learn optimal dropout parameters. However, these methods result in fixed dropout rates that remain static during inference, unable to adapt to new or diverse input distributions encountered after deployment. When the model’s architecture and parameters are inaccessible at inference, or the training dropout setting is unknown, post-hoc dropout methods enable uncertainty estimation without modifying training. Techniques like dropout injection at test time~\cite{ledda2023dropout} and infer-dropout~\cite{mi2022training} apply dropout retrospectively to pre-trained models, offering flexibility in inference, such as widening predictive intervals in safety-critical settings. However, their reliance on fixed dropout rates makes them insensitive to data variability and may be suboptimal in heterogeneous settings like medical applications.

To address these limitations, we propose leveraging information theory to dynamically adjust dropout rates during inference. By viewing dropout as controlled noise injection and each layer as a noisy communication channel~\cite{cover1999elements}, we can quantify the information loss induced by dropout in each layer's feature maps. This perspective allows us to adjust dropout rates per layer and per input instance, aiming to maintain a desired level of information retention while introducing sufficient variability for uncertainty estimation.

In this paper, we introduce \textbf{Rate-In}, an algorithm that dynamically adjusts dropout rates during inference by quantifying the functional information loss in feature maps induced by dropout. Rate-In requires no ground truth labels, making it suitable for deployment in real-world applications where labeled data is unavailable or scarce. By treating each layer's dropout-induced information loss as a controllable parameter, Rate-In adaptively tunes dropout rates to preserve perceptual quality across  tasks and architectures.

Our main contributions are:

\begin{itemize} \item We introduce an information-theoretic framework for adjusting dropout rates during inference, interpreting dropout as controlled noise injection and quantifying its impact on information flow within the network. \item We propose \textbf{Rate-In}, an algorithm that dynamically adjusts dropout rates per layer and per input instance by controlling the information loss induced by dropout, without requiring ground truth labels. \item We demonstrate Rate-In's effectiveness through extensive empirical evaluations on synthetic data and real-world medical imaging tasks, showing improved uncertainty estimation and calibration over fixed or heuristic dropout rates, without compromising predictive performance. \end{itemize}

The rest of the paper is organized as follows: In Section~\ref{sec:background}, we provide background on uncertainty estimation in neural networks, discuss dropout and its limitations, and explore the application of information theory to deep learning. In Section~\ref{sec:rate_in}, we detail the Rate-In algorithm and its implementation. Section~\ref{sec:experiments} presents experimental results demonstrating the benefits of Rate-In in uncertainty estimation tasks. Finally, in Section~\ref{sec:summary}, we discuss the implications of our findings and potential directions for future work.
\section{Background}
\label{sec:background}

This section provides the necessary background for our method. We first discuss the role of uncertainty estimation in neural networks and the existing approaches to achieve it. Next, we examine Monte Carlo (MC) Dropout as a practical method for uncertainty estimation, highlighting its limitations due to fixed dropout rates during inference. Finally, we introduce the application of information theory in deep learning, particularly how it can inform the adaptive adjustment of dropout rates to improve uncertainty estimation.

\subsection{Uncertainty Estimation in Neural Networks}

Deep neural networks have revolutionized numerous fields due to their remarkable predictive capabilities. However, their deterministic nature often prevents them from expressing uncertainty about their predictions~\cite{Gal2016UncertaintyID}. In risk-sensitive applications such as medical diagnosis, autonomous driving, and financial forecasting, understanding the confidence of a model's predictions is as crucial as the predictions themselves~\cite{begoli2019need}. Reliable uncertainty estimation allows practitioners to make informed decisions, identify cases where the model may be less reliable, and potentially defer to human experts when necessary.

Uncertainty in neural networks can be broadly categorized into two types~\cite{kendall2017uncertainties}:

\begin{itemize}
   \item \textbf{Aleatoric Uncertainty}: Arises from inherent data noise and represents the uncertainty due to the stochastic nature of the data generation process. It is irreducible and cannot be eliminated even with more data.
   \item \textbf{Epistemic Uncertainty}: Stems from model uncertainty, often due to limited data or model capacity, and can be reduced with more data or better modeling.
\end{itemize}

\noindent Accurately estimating both types of uncertainty is essential for building reliable models. Various methods have been proposed for uncertainty estimation in neural networks; below are key representative approaches:

\begin{itemize}
   \item \textbf{Bayesian Neural Networks (BNNs)}: Introduce probability distributions over network weights~\cite{blundell2015weight}, allowing the model to capture uncertainty through posterior distributions. However, BNNs are often computationally expensive and challenging to scale.
   \item \textbf{Ensemble Methods}: Train multiple models and aggregate their predictions~\cite{lakshminarayanan2017simple}. Ensembles can capture both types of uncertainty but require significant computational resources.

\item \textbf{Test-time Data Augmentation}: Repeated input perturbation during inference (e.g., rotation, blurring) to approximate a predictive distribution. Effective in domains where transformations are guided by domain knowledge~\cite{mi2022training}.

\item \textbf{Noise Injection Techniques}:  Introduces controlled noise to the model (e.g., Gaussian noise added to weights) to assess sensitivity beyond input-space augmentation~\cite{mi2022training}.

\item \textbf{Monte-Carlo (MC) Techniques}: Use stochastic sampling to approximate Bayesian uncertainty. MC Dropout applies dropout at inference to sample the model’s parameter space~\cite{gal2016dropout}. MC Batch Normalization perturbs batch statistics by sampling batches from the training data during inference, making it effective when training data is accessible~\cite{teye2018bayesian}. Masksembles employs fixed binary masks with controlled sparsity and overlap, balancing MC Dropout’s randomness with Deep Ensembles’ independence, but without posterior sampling~\cite{durasov2021masksembles}. MC Layer Normalization introduces stochasticity by subsampling features (e.g., via fixed-rate dropout) and computing layer norm statistics on the sampled feature space~\cite{frick2024mc}. 
   
\end{itemize}

\subsection{MC Dropout for Uncertainty Estimation}

Monte Carlo Dropout leverages the dropout regularization technique~\cite{srivastava2014dropout} for uncertainty estimation by performing multiple stochastic forward passes through the network with dropout activated during inference~\cite{gal2016dropout}. This approach approximates a Bayesian posterior over the model parameters, allowing for the estimation of predictive uncertainty.

Given an input $\mathbf{x}$ and a neural network with parameters $\mathbf{W}$, the predictive distribution $\mathbf{y}$ is approximated by averaging the predictions over $T$ stochastic forward passes:
\begin{equation}
   p(\mathbf{y} | \mathbf{x}) \approx \frac{1}{T} \sum_{t=1}^{T} p(\mathbf{y} | \mathbf{x}, \mathbf{W}_t),
\end{equation}
where $\mathbf{W}_t$ represents the parameters of the network after applying dropout at iteration $t$. The variability in the predictions reflects the model's uncertainty.

\subsection{Limitations of Fixed Dropout Rates}
While MC Dropout is a practical method for uncertainty estimation, it typically applies fixed dropout rates across all layers and inputs during inference~\cite{gal2016dropout}. This approach has limitations:

\begin{itemize}
   \item \textbf{Suboptimal Uncertainty Estimation}: Fixed dropout rates may not adequately capture the predictive variance needed for accurate uncertainty estimation~\cite{osband2016risk}. Some inputs may require higher dropout rates to reflect uncertainty, while others may suffer from excessive noise.
   \item \textbf{Potential Performance Degradation}: Applying the same dropout rate uniformly can introduce unnecessary noise in some layers, potentially affecting model performance~\cite{srivastava2014dropout}. Critical features may be disrupted, adversely impacting prediction accuracy.
   \item \textbf{Lack of Adaptability}: Fixed rates ignore the varying complexity of inputs or the differing sensitivities of layers within the network. This rigidity limits the model's ability to adapt to new or diverse data distributions.
\end{itemize}

\noindent These limitations highlight the need for methods that adapt dropout rates dynamically during inference to improve uncertainty estimation without compromising performance.

\subsection{Information Theory in Deep Learning}

Information theory provides a powerful framework for quantifying the amount of information transmitted through a system~\cite{cover1999elements}. In the context of deep learning, information-theoretic concepts have been used to understand and improve neural networks in several ways:

\begin{itemize}
   \item \textbf{Information Bottleneck Principle}: Proposes that neural network layers aim to compress input data while preserving relevant information for the output~\cite{shwartz2017opening}. It has been used to analyze learning dynamics and generalization.
   \item \textbf{Mutual Information Analysis}: Estimating mutual information between inputs, hidden layers, and outputs helps in understanding how information is processed and transformed within the network~\cite{shwartz2022information, belghazi2018mine}.
   \item \textbf{Regularization Techniques}: Information-theoretic regularization methods, such as information dropout~\cite{achille2018information}, have been introduced to improve robustness and generalization by controlling the flow of information during training.
\end{itemize}

\noindent However, most of these applications focus on the training of neural networks, using information theory to guide learning and improve representations. The potential of information theory to drive adaptive mechanisms during inference, particularly for uncertainty estimation, remains underexplored.

\subsection{Applying Information Theory to Dropout}

Interpreting dropout through information theory allows us to view it as introducing controlled noise into the network, effectively treating each layer as a \emph{noisy communication channel}~\cite{cover1999elements}. This perspective provides valuable insights into dropout's impact information flow within the network:

\begin{itemize}
   \item \textbf{Dropout as Noise Injection}: Dropout reduces mutual information between a layer's input and output by introducing random perturbations~\cite{srivastava2014dropout, achille2018information}. The noise can prevent overfitting by discouraging the network from relying too heavily on any single neuron.
   \item \textbf{Information Loss Quantification}: The amount of information lost due to dropout can be quantified using measures like entropy and mutual information~\cite{chelombiev2019adaptive}. This quantification enables us to assess the impact of dropout on the network's representations.
   \item \textbf{Layer Sensitivity}: Different layers may exhibit varying sensitivities to information loss induced by dropout, depending on their role and the nature of their activations. Early layers may be more robust to noise, while later layers may be more sensitive.
\end{itemize}

\noindent These insights suggest that by controlling the information loss induced by dropout, we can adaptively adjust dropout rates to balance the trade-off between introducing variability for uncertainty estimation and maintaining sufficient information for accurate predictions.
\section{Rate-In: Adaptive Dropout at Inference}
\label{sec:rate_in}

In this section, we present \textbf{Rate-In}, an algorithm that dynamically adjusts dropout rates during inference by quantifying the information loss in each layer's feature maps. Treating dropout as controlled noise injection and leveraging information-theoretic principles, Rate-In adapts dropout rates per layer and per input instance, enhancing uncertainty estimation without compromising predictive performance.

\subsection{Reinterpreting Dropout as Noise Injection}

Traditionally, dropout serves as a regularization technique that randomly drops units or connections during training~\cite{srivastava2014dropout, wan2013regularization}. For uncertainty estimation during inference, we reinterpret dropout as a form of noise injection into the neural signal. This perspective allows us to analyze dropout's impact on information flow within the network.

Treating each layer as a communication channel~\cite{cover1999elements}, dropout introduces noise into this channel, reducing the mutual information between the layer's input and output. The extent of information loss depends on the dropout rate and the characteristics of the activations in that layer.

\subsection{Quantifying Information Loss}

To dynamically adjust dropout rates, we need to quantify the information loss induced by dropout in each layer. We use \emph{mutual information} (MI), which measures the shared information between two random variables~\cite{cover1999elements}.

Given a layer with input activations $\mathbf{h}_\text{in}$ and output activations $\mathbf{h}_\text{out}$, the mutual information $I(\mathbf{h}_\text{in}; \mathbf{h}_\text{out})$ quantifies how much information about $\mathbf{h}_\text{in}$ is preserved in $\mathbf{h}_\text{out}$. Dropout reduces this mutual information due to the introduced randomness.

Directly computing MI in high-dimensional spaces is challenging. To address this, we use efficient estimation methods, such as the \emph{adaptive binning estimator}~\cite{chelombiev2019adaptive}, which approximates MI using entropy estimations with adaptive data binning.

\subsection{The Rate-In Algorithm}

\begin{figure}[htbp]
\includegraphics[height=3.70cm]{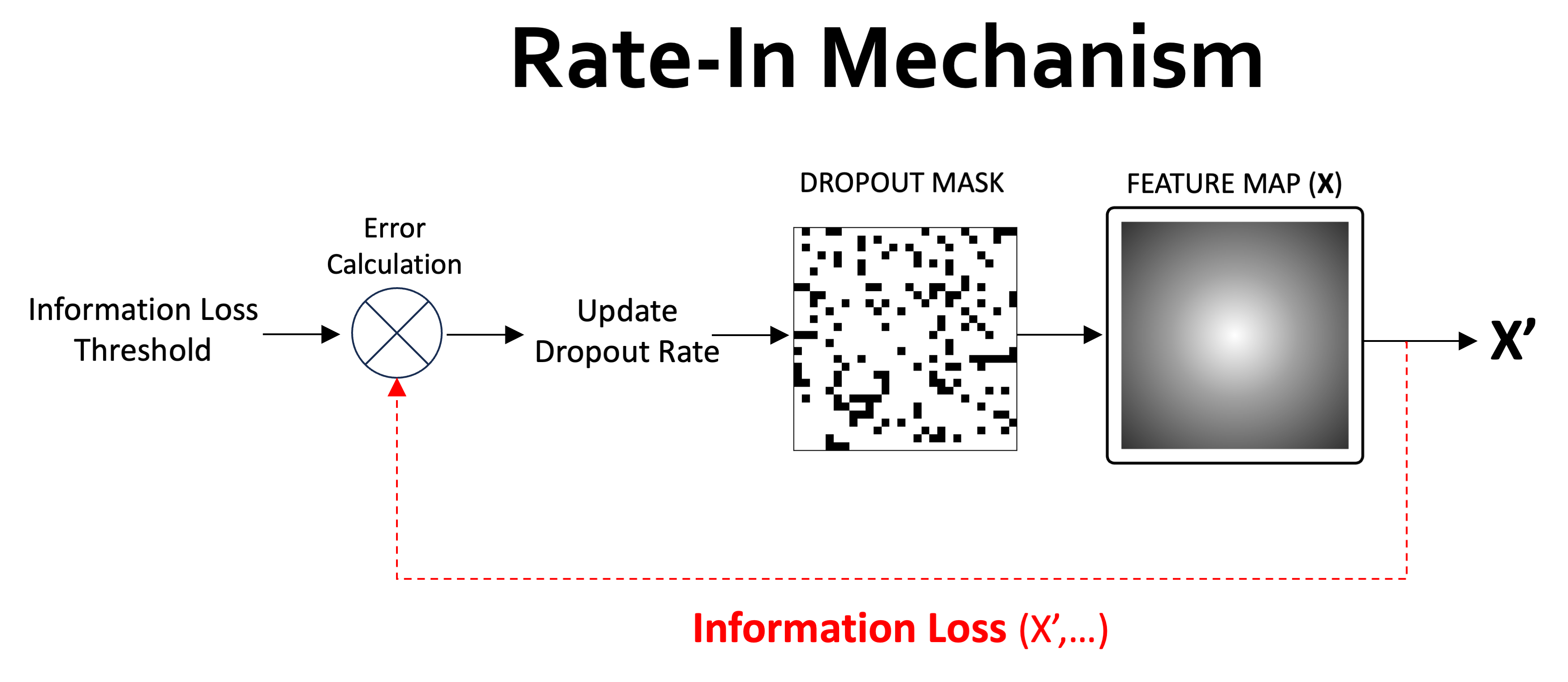} 
\caption{\small \textbf{Rate-In dynamically optimizes dropout rates through an information-theoretic feedback loop:} (i) apply a dropout mask to the feature map; (ii) assess information integrity; (iii) calculate the difference from the target information level; (iv) update the dropout rate accordingly.} \label{fig:feedback_loop_schematic} 

\end{figure}

Rate-In adjusts the dropout rate for each layer during inference by maintaining the information loss below a predefined threshold. The algorithm operates sequentially during the forward pass, adjusting dropout rates per layer and per input instance (Figure \ref{fig:feedback_loop_schematic} and \cref{alg:ratein}).

\subsubsection{Algorithm Details}

\begin{algorithm}[t] \caption{Rate-In Algorithm} \label{alg:ratein} \begin{algorithmic}[1] \Require{Trained neural network $\mathcal{F}$ with $L$ layers, input $\mathbf{x}$, initial dropout rates $p^{(0)}$, information loss threshold $\epsilon$, maximum iterations $N_{\text{max}}$} \Ensure{Adjusted dropout rates $p$, output prediction $\hat{\mathbf{y}}$} \State Initialize $p \gets p^{(0)}$ \For{$l = 1$ to $L$} \State $n \gets 0$ 
\Repeat 
    \State Apply dropout with rate $p_l$ to layer $l$ 
    \State Compute activations $\mathbf{h}^{(l)}_{\text{in}}$ and $\mathbf{h}^{(l)}_{\text{out}}$ 
    \State Estimate mutual information $I(\mathbf{h}^{(l)}_{\text{in}}; \mathbf{h}^{(l)}_{\text{out}})$ 
    \State Compute information loss $\Delta I_l = I^{(l)}_{\text{full}} - I(\mathbf{h}^{(l)}_{\text{in}}; \mathbf{h}^{(l)}_{\text{out}})$ \If{$\Delta I_l > \epsilon$} 
        \State Decrease $p_l$ 
    \Else \State Increase $p_l$ 
    \EndIf \State $n \gets n + 1$ \Until{$|\Delta I_l - \epsilon| < \delta$ \textbf{or} $n \geq N{\text{max}}$} \EndFor \State Perform forward pass with adjusted dropout to obtain $\hat{\mathbf{y}}$ \State \Return $p$, $\hat{\mathbf{y}}$ \end{algorithmic} \end{algorithm}

\paragraph{Initialization} We start with initial dropout rates $p^{(0)}$, which can be uniform across layers or based on prior knowledge. We define an information loss threshold $\epsilon$ specifying the maximum allowable mutual information loss per layer.

\paragraph{Layer-wise Adjustment}
For each layer $l$, we perform the following steps:

\begin{enumerate}
   \item \textbf{Apply Dropout}: Apply dropout with the current rate $p_l$ to the activations in layer $l$.
   \item \textbf{Compute Activations}: Obtain the input and output activations $\mathbf{h}^{(l)}_{\text{in}}$ and $\mathbf{h}^{(l)}_{\text{out}}$ for layer $l$.
   \item \textbf{Estimate Mutual Information}: Estimate the mutual information $I(\mathbf{h}^{(l)}_{\text{in}}; \mathbf{h}^{(l)}_{\text{out}})$ using an appropriate estimator.
   \item \textbf{Compute Information Loss}: Calculate the information loss $\Delta I_l = I^{(l)}_{\text{full}} - I(\mathbf{h}^{(l)}_{\text{in}}; \mathbf{h}^{(l)}_{\text{out}})$, where $I^{(l)}_{\text{full}}$ is the mutual information without dropout.
   \item \textbf{Adjust Dropout Rate}: If $\Delta I_l > \epsilon$, decrease $p_l$ to reduce information loss; else, increase $p_l$ to enhance variability.
   \item \textbf{Convergence Check}: Repeat the adjustment until the information loss is within an acceptable range or the maximum number of iterations is reached.
\end{enumerate}

\paragraph{Forward Pass} After adjusting dropout rates for all layers, perform a final forward pass with adjusted rates to obtain $\hat{\mathbf{y}}$.

\subsubsection{Implementation Considerations}
\label{sec:implementation_considerations}
\paragraph{Estimating Mutual Information} To make MI estimation tractable in high-dimensional spaces, we can:

\begin{itemize} \item Use \textbf{dimensionality reduction} to project activations to lower dimensions. \item Employ \textbf{non-parametric estimators} like k-nearest neighbors or kernel density estimators. \item Approximate MI using \textbf{simpler metrics} such as correlation or variance when appropriate. \end{itemize}

\paragraph{Optimization Strategy} Adjusting dropout rates can be formulated as an optimization problem. Simple strategies like gradient descent or rule-based updates (e.g., fixed step sizes) can achieve the target information loss threshold.

\paragraph{Computational Overhead} To mitigate computations:

\begin{itemize} 
\item Limit the number of iterations per layer ($N_{\text{max}}$).
\item Share computations across similar inputs or batchs. 
\item Precompute or approximate certain quantities. 
\end{itemize}

\subsection{Discussion of the Rate-In Approach}

The Rate-In algorithm offers several advantages:

\begin{itemize}
   \item \textbf{Adaptive Dropout Rates}: By adjusting dropout rates per layer and input instance, Rate-In accommodates varying layers sensitivities and input complexities.
   \item \textbf{Improved Uncertainty Estimation}: By controlling the information loss, Rate-In ensures that sufficient variability is introduced for uncertainty estimation without degrading the predictive performance.
   \item \textbf{No Need for Ground Truth Labels}: Rate-In operates during inference without labeled data, making it applicable in deployment scenarios where labels are unavailable.
\end{itemize}

However, there are also challenges:

\begin{itemize}
   \item \textbf{Computational Complexity}: The need to estimate MI and adjust dropout rates can increase inference time.
   \item \textbf{Estimation Accuracy}: Accurately estimating MI is non-trivial, especially in high-dimensional spaces.
   \item \textbf{Hyperparameter Selection}: Choosing appropriate values for the information loss threshold $\epsilon$ and other hyperparameters requires careful consideration.
\end{itemize}

\noindent Next, we present experimental results demonstrating Rate-In's effectiveness across various tasks and architectures.
\section{Experiments}
\label{sec:experiments}

In this section, we evaluate the effectiveness of the proposed \textbf{Rate-In} algorithm for predictive uncertainty estimation through Monte Carlo (MC) simulations. We test it on synthetic data and real-world medical imaging tasks with state-of-the-art pre-trained neural network architectures. Our evaluation focuses on three main objectives:

\begin{enumerate}[label=(\roman*)] \item \textbf{Effectiveness of Rate-In}: Assess Rate-In's ability to improve uncertainty estimation and calibration compared to fixed or heuristic dropout rates. \item \textbf{Impact on Predictive Performance}: Determine whether dynamically adjusting dropout rates at inference time affects predictive accuracy. \item \textbf{Adaptability across Tasks and Architectures}: Demonstrate the applicability of Rate-In across different tasks and neural network architectures. \end{enumerate}

\subsection{Experimental Setup}
\label{sec:exp_setup}

We evaluate our method on four categories of tasks:

\vspace{-12pt}
\paragraph{Synthetic Regression}

To validate our method in a controlled setting, we generate a one-dimensional regression dataset with varying levels of noise. The target variable is defined as $y = \sin(x) + \epsilon$, where $x$ is uniformly sampled from $[-3, 3]$ and $\epsilon \sim \mathcal{N}(0, \sigma^2)$. We create datasets with five different noise levels by setting $\sigma \in \{0.1, 0.2, 0.3, 0.4, 0.5\}$. Each dataset contains 100 training and test instances.

\vspace{-12pt}
\paragraph{Medical Image Classification}

For real-world classification tasks, we use datasets from MedMNIST v2~\cite{yang2023medmnist}:
\begin{itemize} 
    \item \textbf{PathMNIST}: 107,180 histopathology images for colorectal cancer tissue classification (9 classes).
    \item \textbf{BloodMNIST}: 17,092 blood cell microscope images for peripheral blood cell classification (8 classes). 
    \item \textbf{TissueMNIST}: 236,386 kidney cortex microscope images for kidney cortex cell type classification (8 classes). 
    \end{itemize}

\vspace{-12pt}
\paragraph{Medical Image Segmentation}

We select two segmentation tasks involving different imaging modalities:

\begin{itemize} 
    \item \textbf{Prostate MRI Segmentation}~\cite{antonelli2022medical}: Segmenting the peripheral and transitional zones in prostate MRI scans. 
    \item \textbf{Liver Tumor CT Segmentation}~\cite{antonelli2022medical}: Segmenting liver and tumor regions in CT scans. 
\end{itemize}

\vspace{-12pt}
 
\paragraph{Out of Distribution Robustness}
To evaluate robustness to real-world medical imaging artifacts, we used MedMNIST-C~\cite{di2024medmnist}, a dataset with controlled task- and modality-specific corruptions, similar to ImageNet-C~\cite{hendrycks2019benchmarking}. These include common artifacts such as air bubbles, defocus, and motion blur, applied at varying severity levels. We tested the classification datasets in their corrupted versions: \textbf{PathMNIST-C}, \textbf{BloodMNIST-C}, and \textbf{TissueMNIST-C}.
\\

\noindent We compare Rate-In with the following \textbf{baseline methods}:

\begin{itemize} 
    \item \textbf{Fixed Dropout} \cite{srivastava2014dropout}: Applying a constant dropout rate across all layers during MC iterations. 
    \item \textbf{Scheduled Dropout} \cite{rennie2014annealed}: Linearly decreasing the dropout rate over MC iterations. 
    \item \textbf{Activation-Based Dropout}: Adjusting dropout rates based on the coefficient of variation (CoV) of layer activations, inspired by adaptive dropout methods~\cite{ba2013adaptive}.
\end{itemize}

\noindent Formal definitions of these dropout strategies are provided in Appendix A Section ~\ref{sec:suppl_baselines}.

\subsubsection{Implementation Details}

To ensure a fair comparison, Rate-In's information loss threshold ($\epsilon$) was set equal to the dropout rate ($p^{(0)}$) of other benchmarks, emphasizing the conversion of impulse noise to contextual noise. With $p_0=p$, all methods start equally, but Rate-In dynamically adjusts dropout rates based on layer-wise redundancy. $\delta$ was set to 0.01 and $N_{\text{max}}$ to 30 based on preliminary experiments.

For information loss estimation, we use the adaptive binning estimator~\cite{chelombiev2019adaptive} to efficiently approximate mutual information in high-dimensional feature spaces. We also assess feature preservation using the structural similarity index (SSIM)~\cite{wang2004image} (See Appendix B for details and results). MC simulations use 30 forward passes per input, with dropout rates dynamically adjusted based on measured information loss. We employ the following network architectures:

\begin{itemize} 
\item \textbf{Synthetic Regression}: A fully connected network with two hidden layers of 50 units each and ReLU activations. \item \textbf{Classification Tasks}: ResNet-18~\cite{he2016deep} pre-trained on MedMNIST datasets, with dropout layers added after each residual block's ReLU activation. 
\item \textbf{Segmentation Tasks}: nnU-Net~\cite{isensee2021nnu}, a U-Net-based architecture with dropout after each block. 
\end{itemize}

\paragraph{Evaluation Metrics}

For predictive performance, we use Mean Squared Error (MSE) for regression, Accuracy (ACC) for classification, and Dice Similarity Coefficient (DSC) for segmentation. For uncertainty estimation, we employ Expected Calibration Error (ECE)~\cite{guo2017calibration}, Area Under the Accuracy-Rejection Curve (AUARC)~\cite{geifman2018bias}, and Boundary Uncertainty Coverage (BUC)~\cite{yue2024boundary}. We also provide qualitative analyses through uncertainty maps. Full metric formulations are in Appendix A Section ~\ref{sec:suppl_eval_metrics}.

\paragraph{Computational Complexity}

We analyze the theoretical complexity of Rate-In optimization and validate it empirically. Sensitivity to key parameters - initial dropout rate, noise level, information loss threshold, and dataset size, is studied on synthetic data, while inference overhead is assessed in real-world classification and segmentation tasks.

\paragraph{Reproducibility and Ablation Studies} 
Code and implementation details are in the supplementary material. \textbf{Appendix A} covers methodology, training protocols, and evaluation. \textbf{Appendix B} includes additional results, ablations, and complexity analysis. \textbf{Appendix C} discusses Rate-In’s theoretical properties and implications.

\subsection{Results}

\subsubsection{Synthetic Data Analysis}

\begin{figure}[ht]
    \centering
    \begin{subfigure}{0.49\columnwidth}
        \includegraphics[width=\linewidth]{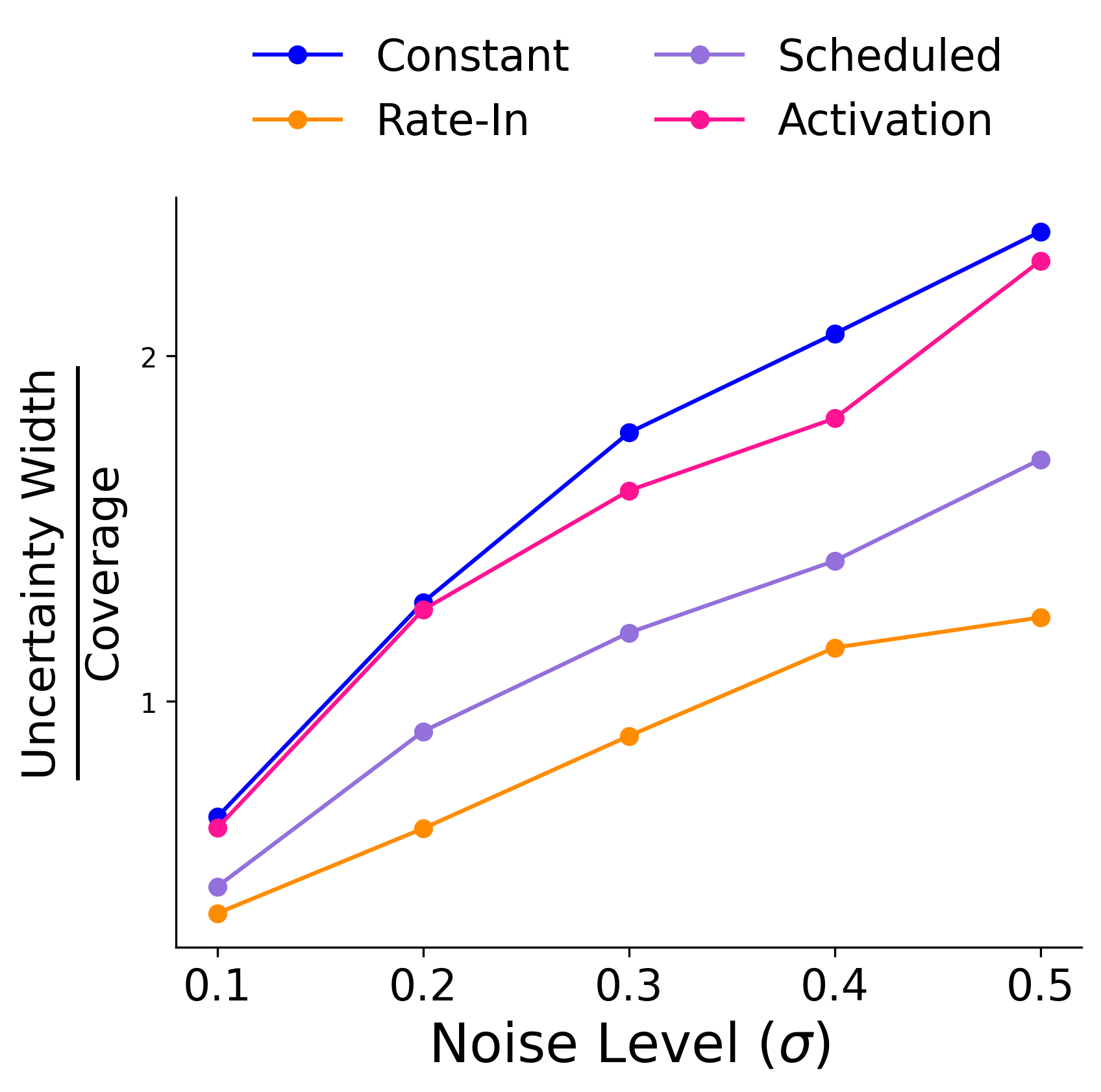}
        \caption{\scriptsize Increasing noise level}
        \label{fig:synthetic_data_efficiency_noise}
    \end{subfigure}
    \hfill
    \begin{subfigure}{0.49\columnwidth}
        \includegraphics[width=\linewidth]{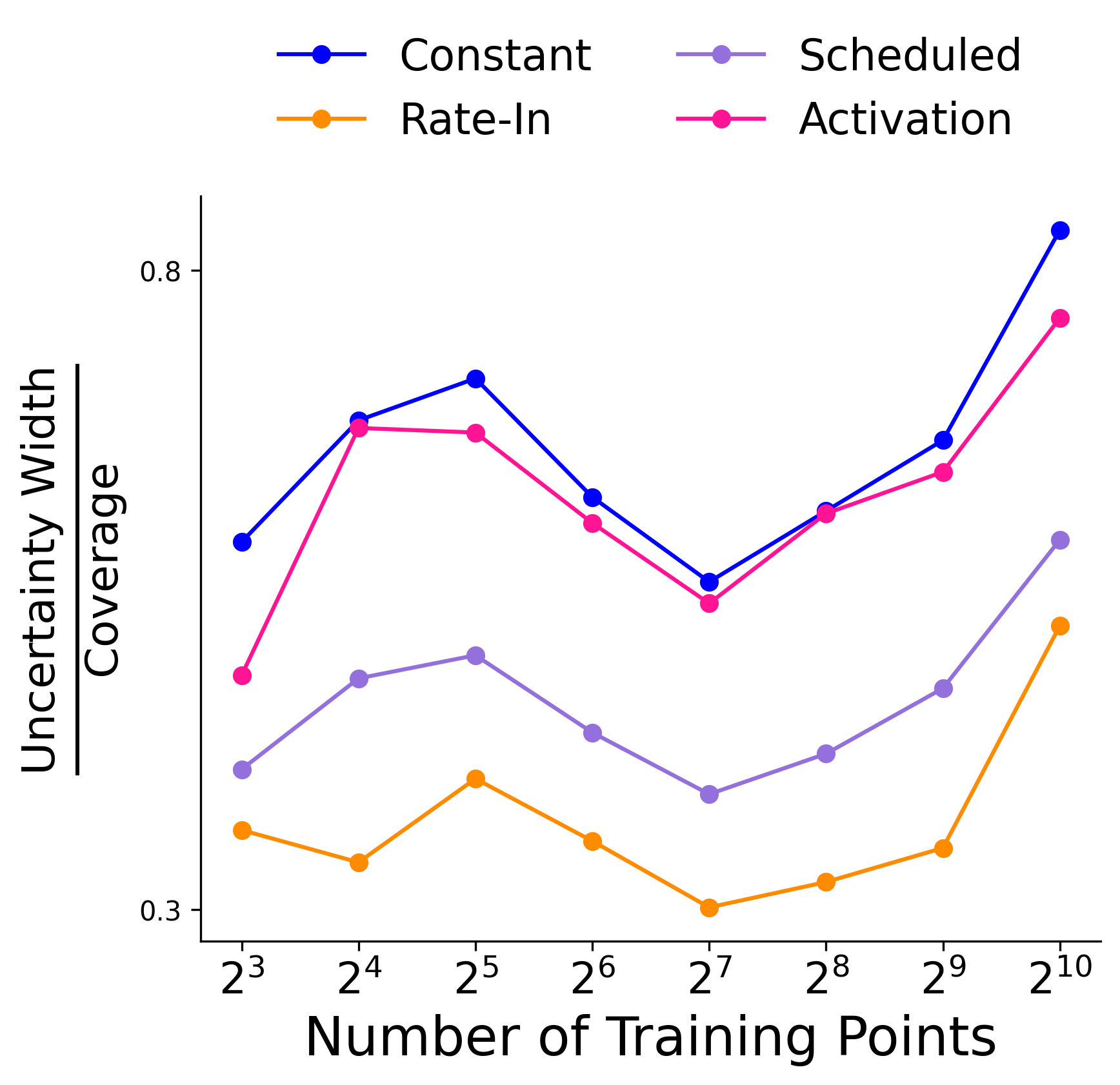}
        \caption{\scriptsize  Increasing training points}
        \label{fig:synthetic_data_efficiency_training}
    \end{subfigure}
    \hfill
    \vspace{-5pt}
    \caption{\small \textbf{Rate-In is more efficient in uncertainty estimation}. The ratio of uncertainty interval width to 95\% coverage. Lower ratios indicate greater efficiency. (a) Varying noise levels; (b) Varying number of training points at fixed noise level. $(\sigma = 0.01)$.}
    \label{fig:synthetic_data_efficiency}
    \vspace{-5pt}
\end{figure}

We first evaluate Rate-In on the synthetic regression task to validate its performance under controlled conditions. \cref{fig:synthetic_data_efficiency_noise} shows the efficiency of uncertainty estimation across increasing noise levels ($\sigma \in \{0.1, 0.2, 0.3, 0.4, 0.5\}$) with a fixed training set size ($N=100$). Efficiency is measured as the ratio of interval width to empirical coverage; lower ratios indicate more precise uncertainty estimates without compromising reliability. Rate-In maintains a consistently lower ratio across all noise levels compared to baselines. 

\cref{fig:synthetic_data_efficiency_training} shows the impact of training set size under a fixed noise level ($\sigma = 0.01$). Rate-In outperforms other methods across all sample sizes, achieving more efficient uncertainty estimation.The results highlight Rate-In's ability to adapt dropout rates based on local uncertainty characteristics, leading to more efficient and reliable estimates than other dropout baselines. The method is particularly advantageous in scenarios with heteroscedastic noise and limited training data. Sensitivity analysis on training size, noise, and convergence is in Appendix B, Sections ~\ref{supp:synthetic_data} and ~\ref{supp:complexity_analysis}.

\subsubsection{Classification Tasks}
Table \ref{tab:classification} presents performance metrics for each classification task and dropout policy. Rate-In consistently outperforms baselines in accuracy and uncertainty calibration. Specifically, for the PathMNIST dataset, Rate-In maintains high performance across varying information loss levels. At a 0.05 dropout rate, Rate-In preserves full model accuracy for both TissueMNIST and PathMNIST datasets, while other methods degrade significantly.

The three medical datasets exhibit distinct patterns of robustness to dropout. BloodMNIST retains high performance even under significant dropout, while TissueMNIST shows greater sensitivity to information loss, and PathMNIST demonstrates intermediate resilience. These differences highlight the importance of adaptive dropout strategies across different medical imaging applications.
 
\begin{table}[htbp]
\renewcommand{\arraystretch}{0.8}
\setlength{\abovecaptionskip}{0pt}
\setlength{\belowcaptionskip}{0pt}
\footnotesize
\centering
\caption{\small \textbf{Rate-In preserves near-original model accuracy even at high dropout rates, while baseline methods degrade significantly.} 
    Classification performance across medical imaging tasks is reported as Accuracy (AUARC). Rate-In consistently outperforms traditional dropout methods, particularly at higher dropout rates. Higher AUARC values indicate better uncertainty estimation.}\label{tab:classification}%
\setlength{\tabcolsep}{5.0pt}%
\begin{tabular}{@{}p{1.4cm}p{1.0cm}c ccc@{}}
\toprule
\textbf{Dataset} & \textbf{Method} & \textbf{Accuracy} & \multicolumn{3}{c}{\textbf{Accuracy (AUARC)}} \\
\cmidrule(lr){4-6}
& & \scriptsize{Full Model} & $p\text{,}\epsilon{\text{=}}0.05$ & $p\text{,}\epsilon{\text{=}}0.10$ & $p\text{,}\epsilon{\text{=}}0.20$ \\
\midrule[0.3pt]
\multirow{4}{*}[-1pt]{\textit{TissueMNIST}} 
& \textbf{Rate-In} & \multirow{4}{*}{.68} & \textbf{.68} (\textbf{.84}) & \textbf{.64} (\textbf{.80}) & \textbf{.48} (\textbf{.59}) \\
& Constant   & & .53 (.67) & .31 (.37) & .09 (.25) \\
& Scheduled  & & .56 (.70) & .36 (.47) & .13 (.19) \\
& Activation & & .54 (.67) & .32 (.40) & .10 (.24) \\
\cmidrule(lr){1-6}
\multirow{4}{*}[-1pt]{\textit{PathMNIST}} 
& \textbf{Rate-In} & \multirow{4}{*}{.90} & \textbf{.90} (\textbf{.98}) & \textbf{.88} (\textbf{.97}) & \textbf{.75} (\textbf{.91}) \\
& Constant   & & .87 (.96) & .75 (.91) & .41 (.56) \\
& Scheduled  & & .88 (.97) & .78 (.93) & .47 (.65) \\
& Activation & & .87 (.96) & .76 (.92) & .44 (.60) \\
\cmidrule(lr){1-6}
\multirow{4}{*}[-1pt]{\textit{BloodMNIST}} 
& \textbf{Rate-In} & \multirow{4}{*}{.97} & \textbf{.96} (\textbf{.99}) & \textbf{.93} (\textbf{.99}) & \textbf{.82} (\textbf{.92}) \\
& Constant   & & .95 (\textbf{.99}) & .91 (.98) & .58 (.75) \\
& Scheduled  & & .95 (\textbf{.99}) & .92 (.98) & .75 (.86) \\
& Activation & & \textbf{.96} (\textbf{.99}) & \textbf{.93} (\textbf{.99}) & .81 (\textbf{.92}) \\
\bottomrule
\end{tabular}
\vspace{-15pt}
\end{table}

\subsubsection{Segmentation Tasks}

\begin{table*}[t]
\renewcommand{\arraystretch}{0.6}
\setlength{\abovecaptionskip}{0pt}
\setlength{\belowcaptionskip}{0pt}
\centering
\footnotesize
\caption{\textbf{Rate-In achieves superior uncertainty estimation across all anatomical zones while preserving or improving segmentation accuracy compared to top benchmarks.} 
    Comprehensive MRI and CT evaluations show DSC, ECE, and BUC across varying dropout rates. 
    Rate-In consistently improves uncertainty metrics (lower ECE, higher BUC) while maintaining competitive DSC, excelling in challenging regions like the transitional zone and tumor boundaries.  Results for all dropout methods are in  Appendix B, Table~\ref{tab:suppl_segmentation}.}\label{tab:results}\label{tab:segmentation}%
\setlength{\tabcolsep}{3.5pt}%
\begin{tabular}{@{}lllcccccccccc@{}}
\toprule
\multirow{2}{*}{Zone} & \multirow{2}{*}{Modality} & \multirow{2}{*}{Method} & DSC & \multicolumn{3}{c}{DSC (Full Model, \% change)} & \multicolumn{3}{c}{ECE $\times 10^{-3}$} & \multicolumn{3}{c}{BUC} \\
\cmidrule(lr){5-7} \cmidrule(lr){8-10} \cmidrule(lr){11-13}
& & & Full Model & $p\text{,}\epsilon{\text{=}}0.01$ & $p\text{,}\epsilon{\text{=}}0.05$ & $p\text{,}\epsilon{\text{=}}0.10$ & $p\text{,}\epsilon{\text{=}}0.01$ & $p\text{,}\epsilon{\text{=}}0.05$ & $p\text{,}\epsilon{\text{=}}0.10$ & $p\text{,}\epsilon{\text{=}}0.01$ & $p\text{,}\epsilon{\text{=}}0.05$ & $p\text{,}\epsilon{\text{=}}0.10$ \\
\midrule
\multirow{2}{*}{Peripheral} & \multirow{2}{*}{MRI} & Rate-In & \multirow{2}{*}{0.682} & +0.29\% & +2.79\% & +2.35\% & 4.60 & \textbf{4.26} & \textbf{4.58} & \textbf{0.67} & \textbf{0.63} & \textbf{0.61} \\
& & Benchmark$^*$ & & +2.05\% & +2.59\% & +2.05\% & \textbf{4.40} & 5.55 & 7.85 & 0.54 & 0.48 & 0.45 \\
\midrule
\multirow{2}{*}{Transitional} & \multirow{2}{*}{MRI} & Rate-In & \multirow{2}{*}{0.892} & -0.34\% & -0.78\% & -0.67\% & \textbf{3.97} & \textbf{4.03} & \textbf{4.86} & \textbf{0.87} & \textbf{0.84} & \textbf{0.80} \\
& & Benchmark$^*$ & & +0.11\% & -0.78\% & -0.78\% & 4.25 & 6.60 & 10.73 & 0.70 & 0.62 & 0.57 \\
\midrule
\multirow{2}{*}{Liver} & \multirow{2}{*}{CT} & Rate-In & \multirow{2}{*}{0.955} & +1.05\% & +1.05\% & +0.94\% & \textbf{6.73} & \textbf{5.54} & \textbf{4.65} & \textbf{0.93} & \textbf{0.93} & \textbf{0.92} \\
& & Benchmark$^*$ & & 0.00\% & 0.00\% & 0.00\% & 6.90 & 5.80 & 5.00 & 0.92 & 0.91 & 0.91 \\
\midrule
\multirow{2}{*}{Tumor} & \multirow{2}{*}{CT} & Rate-In & \multirow{2}{*}{0.579} & +1.21\% & -0.52\% & -2.59\% & \textbf{1.78} & \textbf{1.78} & \textbf{1.88} & \textbf{0.61} & 0.53 & 0.48 \\
& & Benchmark$^*$ & & +0.17\% & -1.90\% & -1.73\% & 1.80 & 1.80 & 1.90 & 0.60 & \textbf{0.55} & \textbf{0.53} \\
\bottomrule
\end{tabular}
\vspace{-10pt}
\end{table*}

Rate-In is highly effective in medical image segmentation, where accurate uncertainty estimation is crucial. As shown in Figure~\ref{fig:dropout_viz}, Rate-In generates uncertainty maps that accurately highlight anatomically challenging regions, especially along tissue boundaries where radiologists face ambiguity. In contrast, constant dropout diffuses uncertainty at higher dropout rates, obscuring true areas of uncertainty.

\noindent Rate-In demonstrates several key advantages:

\begin{itemize} 
    \item \textbf{Anatomical Precision:} Rate-In accurately pinpoints ambiguous regions, particularly along tissue boundaries, enhancing clinical interpretability. Constant dropout yields diffuse patterns that obscure genuine concerns.

    \item \textbf{Dropout Rate Stability:} Rate-In maintains consistent, focused uncertainty estimates across varying dropout rates, unlike conventional methods that degrade with increased rates, ensuring reliable clinical performance.

   \item \textbf{Error Correlation:} Rate-In's high-uncertainty regions strongly align with prediction errors, indicating better calibration and enhancing trust in the model's predictions.
    
   \item \textbf{Clinical Interpretability:} Rate-In's uncertainty maps align with anatomical structures and boundaries, making them more interpretable. It facilitates the integration of statistical uncertainty into clinical decision-making.

\end{itemize}

\noindent These qualitative observations are quantitatively supported by test set performance metrics. Table~\ref{tab:segmentation} summarizes results for each segmentation task and dropout policy. Rate-In consistently outperforms benchmarks in uncertainty metrics such as ECE and BUC across most anatomical zones. For instance, in regions with boundary ambiguity, Rate-In enhances uncertainty quantification at region interfaces, as reflected by higher BUC scores in the Transitional zone. This improved boundary characterization is critical for clinical applications where precise delineation of overlapping anatomical zones affects diagnostic accuracy.

Results show that Rate-In's information-theoretic dropout rates adjustment yields more accurate and clinically meaningful uncertainty estimates. Its ability to retain focused uncertainty while preserving anatomical detail makes it particularly valuable for medical imaging applications where precision and reliability are paramount.

\begin{figure}[ht]
\centering
    \includegraphics[width=0.3\textwidth]{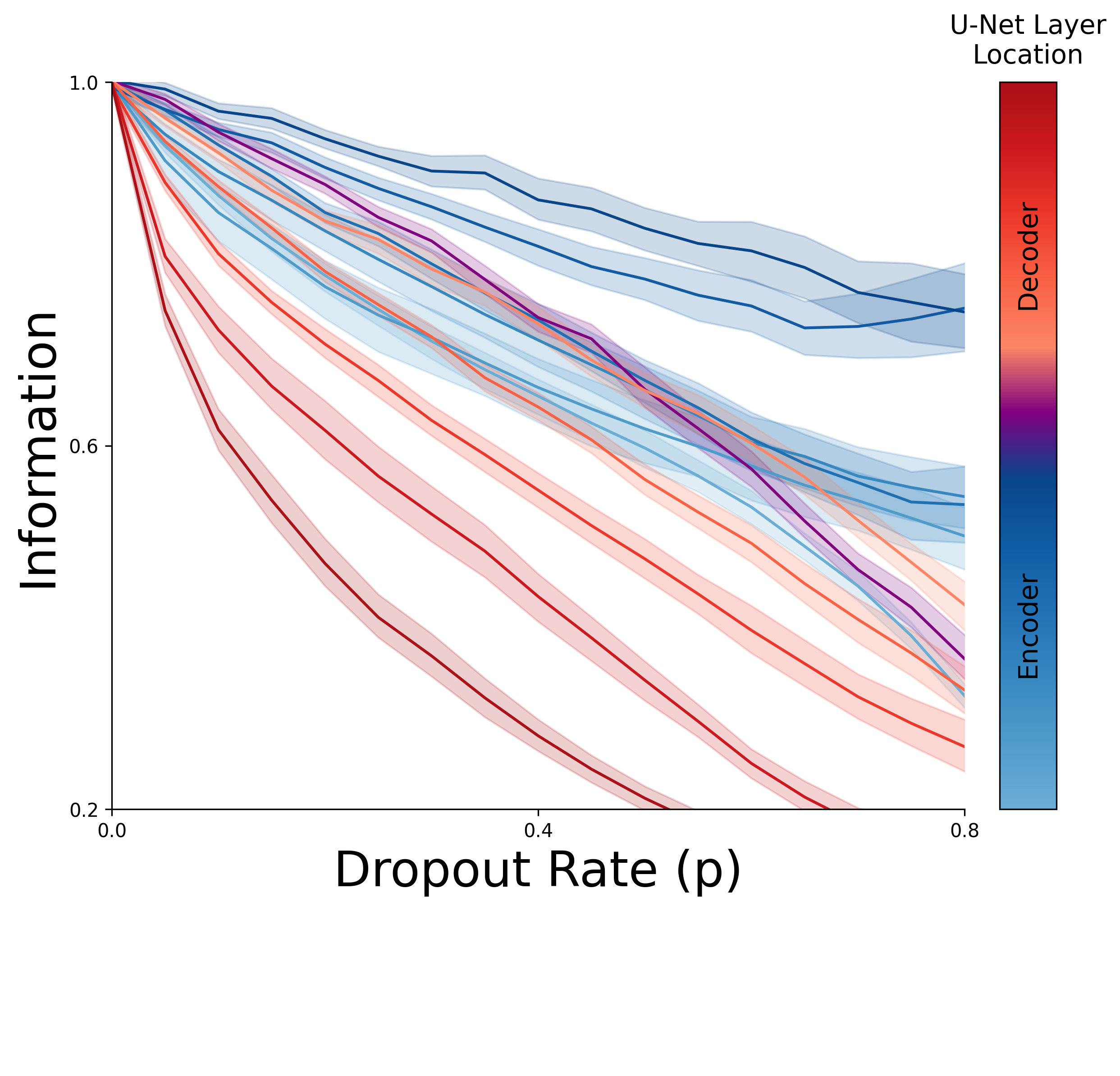}
    \vspace{-1.cm}

 \caption{\textbf{Decoder layers are more sensitive to dropout-induced information loss than encoder layers, highlighting the need for layer-specific dropout rates.} 
    Analysis of information retention across U-Net layers as dropout rates increase. 
    Decoder layers lose more information, while encoder layers remain more stable. 
    This asymmetry supports Rate-In's adaptive approach, assigning lower dropout rates to sensitive decoder layers and higher rates in robust encoder layers for efficient uncertainty estimation.}
\label{fig:seg_sen}
\vspace{-10pt}
\end{figure}

Figure~\ref{fig:seg_sen} illustrates the impact of dropout rates on information loss across network layers. In a pre-trained U-Net model, the decoder layers lose about 40\% of information as dropout rates increase from 5\% to 20\%, while encoder layers remain more stable. This suggests that Rate-In effectively allocates dropout rates to minimize information loss where it is most critical.

\subsection{Out-of-Distribution (OOD) Analysis}
Corrupted test images were generated following the setup and evaluation protocol of \cite{di2024medmnist}. Robustness was assessed per corruption type and overall by averaging performance across types and severity levels. Rate-In improves robustness on MedMNIST-C, maintaining higher accuracy and AUARC across all OOD corruption types compared to the benchmark (Table~\ref{tab:ood}). This improvement is especially notable at higher dropout rates and information loss thresholds, where benchmark accuracy declines more sharply. Rate-In mitigates these drops, demonstrating greater stability across datasets. A detailed breakdown by dropout method and corruption type is in Table~\ref{tab:ood_supp} (Appendix B).

\begin{table}[htbp]
\renewcommand{\arraystretch}{0.6}
\setlength{\abovecaptionskip}{0pt}
\setlength{\belowcaptionskip}{0pt}
\scriptsize
\centering
\caption{\footnotesize \textbf{MedMNIST-C Robustness Results:} ACC(AUARC)}\label{tab:ood}%
\setlength{\tabcolsep}{5.5pt}%
\begin{tabular}{@{}p{1.2cm}p{0.9cm}>{\centering\arraybackslash}p{1.2cm}ccc@{}}
\toprule
\textbf{Dataset} & \textbf{Method} & Full-Model & $p\text{,}\epsilon{\text{=}}0.05$ & $p\text{,}\epsilon{\text{=}}0.10$ & $p\text{,}\epsilon{\text{=}}0.20$ \\
\midrule
\multirow{2}{*}[-1pt]{\textit{TissueMNIST-C}}
& \textbf{Rate-In} & \multirow{2}{*}{.65} & \textbf{.63} (\textbf{.77}) & \textbf{.59} (\textbf{.71}) & \textbf{.45} (\textbf{.48}) \\ 
& Benchmark$^*$ &  & .51 (.60) & .34 (.35) & .12 (.07) \\
\midrule[0.3pt]
\multirow{2}{*}[-1pt]{\textit{PathMNIST-C}} 
& \textbf{Rate-In} & \multirow{2}{*}{.54} & \textbf{.52} (\textbf{.62}) & \textbf{.50} (\textbf{.59}) & \textbf{.39} (\textbf{.49}) \\
& Benchmark$^*$ &  & .49 (.59) & .39 (.49) & .25 (.31) \\
\midrule[0.3pt]
\multirow{2}{*}[-1pt]{\textit{BloodMNIST-C}} 
& \textbf{Rate-In} & \multirow{2}{*}{.77} & \textbf{.77} (\textbf{.87}) & \textbf{.75} (\textbf{.86}) & \textbf{.66} (\textbf{.79}) \\
& Benchmark$^*$ &  & .77 (.87) & .73 (.84) & .59 (.73) \\
\bottomrule
\end{tabular}
\vspace{-10pt}
\end{table}

\subsubsection{Computational Complexity Analysis}  
Rate-In's dropout rate optimization has complexity \(O(N_{\max} \cdot (n + f))\), where \(n\) is the number of input instances, and \(f\) is the complexity of computing information loss. Each iteration involves \(O(n)\) dropout operations and \(O(f)\) calculations, with dropout complexity further decomposing to \(O(n \cdot d)\), where \(d\) is the number of elements per instance. After optimization, inference runs at standard MC-dropout speed. See Appendix B for details.

Empirical results show that classification tasks (2D data, smaller models) complete within 0.70 seconds per instance, while segmentation tasks (3D data, larger models) take a few seconds to 90 seconds (Appendix B, Tables~\ref{tab:rate_in_duration_classification}–\ref{tab:rate_in_duration_segmentation_ssim}).

\section{Summary} \label{sec:summary}

We introduced \textbf{Rate-In}, a novel algorithm that dynamically adjusts dropout rates during inference to enhance predictive uncertainty estimation in neural networks. Traditional methods like Monte Carlo Dropout use fixed dropout rates, which fail to adapt to individual inputs and network layers, leading to suboptimal uncertainty estimates. By treating dropout as controlled noise injection and leveraging information-theoretic principles, Rate-In quantifies the information loss in each layer and adjusts dropout rates accordingly, without requiring additional training or labels.

Our experiments on synthetic data and real-world medical imaging tasks show that Rate-In consistently improves uncertainty calibration and sharpens estimates over fixed or heuristic dropout rates while preserving predictive accuracy. In segmentation, its uncertainty maps highlight anatomically challenging regions, enhancing clinical interpretability. By controlling information loss, Rate-In offers a practical, unsupervised inference-time solution for tuning dropout, making it valuable for risk-sensitive applications requiring precise uncertainty estimates.
{
    \small
    \bibliographystyle{ieeenat_fullname}
    \bibliography{main}
}

\clearpage
\setcounter{page}{1}

\twocolumn[
\begin{@twocolumnfalse}
\begin{center}
    {\LARGE \bfseries Rate-In: Information-Driven Adaptive Dropout Rates for Improved Inference-Time Uncertainty Estimation \par}
    \vspace{0.5em}
    {\large Supplementary Material \par}
\end{center}
\end{@twocolumnfalse}
]

\section{Appendix A: Reproducibility}
\label{suppl:appendix_A}

\subsection{Reproducibility Statement}

The Rate-In algorithm code and implementation examples are available in the \href{https://github.com/code-supplement-25/rate-in/tree/main}{\textbf{GitHub repository}}\footnote{https://github.com/code-supplement-25/rate-in/tree/main}.

\begin{itemize}

\item  \textbf{Environment and Data:} Dependencies are listed in requirements.txt. Datasets (PathMNIST, BloodMNIST, TissueMNIST) are from MedMNIST-V2 ~\cite{yang2023medmnist} and Liver CT and Prostate MRI datasets are from the Medical Segmentation Decathlon ~\cite{antonelli2022medical}. Download and preprocessing instructions are included in the original repositories.

\item \textbf{Training and Evaluation:} Hyperparameters and training schedules are in supplement section ~\ref{sec:suppl_architectures}. Pre-trained models for classification ~\cite{yang2023medmnist} and segmentation ~\cite{isensee2021nnu} are publicly available. Dropout baseline and evaluation metric scripts and formulations are in the repository and supplement section ~\ref{sec:suppl_baselines} and  ~\ref{sec:suppl_eval_metrics}, respectively. 

\item  \textbf{Hardware and Software:} Experiments used Python 3.9.12 and PyTorch 2.3.0 on an NVIDIA GeForce RTX 2080 Ti GPU. The code is hardware-agnostic but runtimes may vary on different setups.

\end{itemize}

\subsection{Experiment Setting Supplement}
\subsubsection{Rate-In Configuration}

To enable direct comparison with standard dropout baseline approaches, we initialized Rate-In's parameters as follows: The initial dropout rate $p_0$ was set equal to the dropout rate $p$ used in the constant dropout benchmark approach. The objective information loss threshold $\epsilon$ was set to $p$, representing similar proportion of information to be preserved. For example, with a constant dropout rate of $p=0.2$, we set $p^{(0)}=0.2$ and $\epsilon=0.2$, targeting 80\% information preservation. We set $\delta$ to 0.01.

We defined information loss at layer $(l)$ as the relative change in mutual information (MI) between input and post-dropout feature maps. Let $I_{\text{drop}}^{(l)}$ and $I_{\text{full}}^{(l)}$ represent MI in dropout and full models at layer $l$:
$$
I_{\text{drop}}^{(l)} = \text{MI}(\mathbf{h}^{(0)}{\text{in}}, \mathbf{h}_\text{drop}^{(l)}{\text{out}})
$$
$$
I_{\text{full}}^{(l)} = \text{MI}(\mathbf{h}^{(0)}{\text{in}}, \mathbf{h}_{\text{full}}^{(l)}{\text{out}})
$$
The information loss at layer $l$ is:
$$\Delta I_l= \frac{I_{\text{drop}}^{(l)} - I_{\text{full}}^{(l)}}{I_{\text{full}}^{(l)}}$$

\paragraph{Network Architectures and MI Calculation:} For synthetic data, we used fully connected networks and calculated mutual information (MI) in two steps: first, computing batch-level MI between the input and each normalized hidden unit using 2D histograms, then averaging these values across all units. For real-world data using CNNs, we first resampled all feature maps to a uniform spatial resolution to maintain spatial consistency. We then calculated MI using adaptive estimators with entropy-equal bins ~\cite{chelombiev2019adaptive}.

\subsubsection{Architectures of Neural Networks Used}
\label{sec:suppl_architectures}
\begin{figure}[H]
\centering
    \includegraphics[width=0.20\textwidth]{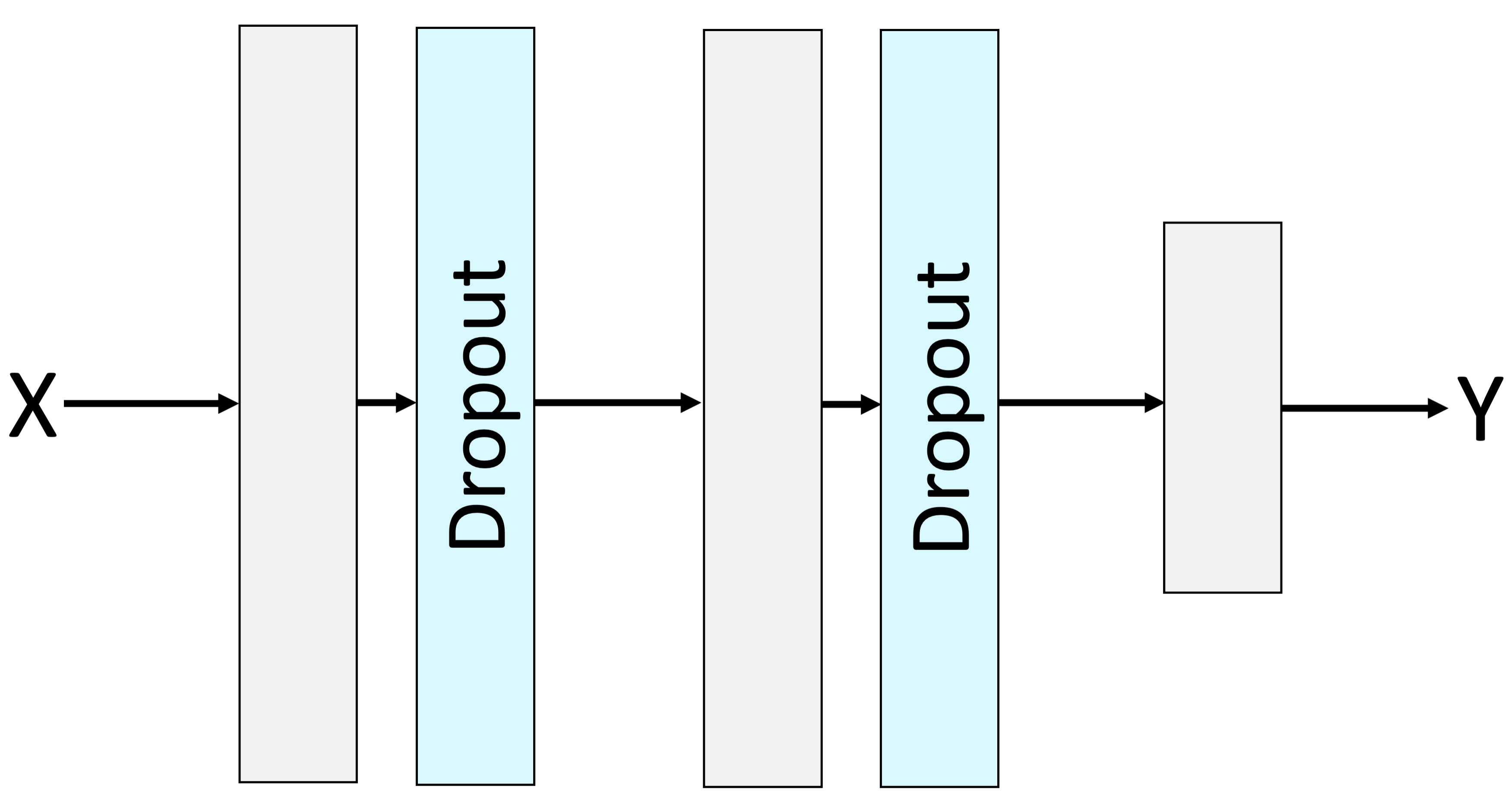}

 \caption{Basic layout of the regression network with dropout layers shown in blue.}
\label{fig:arc_fc}
\vspace{-10pt}
\end{figure}

\paragraph{Synthetic Data:}
The network was implemented in PyTorch with an architecture consisting of three linear layers ([1→50→50→1]) with ReLU activations between hidden layers. Training utilized Adam optimizer (lr=0.01) and mean squared error (MSE) loss over 1000 epochs, processing the entire dataset (N=100) in each iteration. For reproducibility, NumPy and PyTorch random seeds were set to 123. Dropout layers were placed after each hidden layer post-training (Figure ~\ref{fig:arc_fc}).
\vspace{-5pt}

\vspace{-10pt}

\begin{figure}[H]
\centering
    \includegraphics[width=0.25\textwidth]{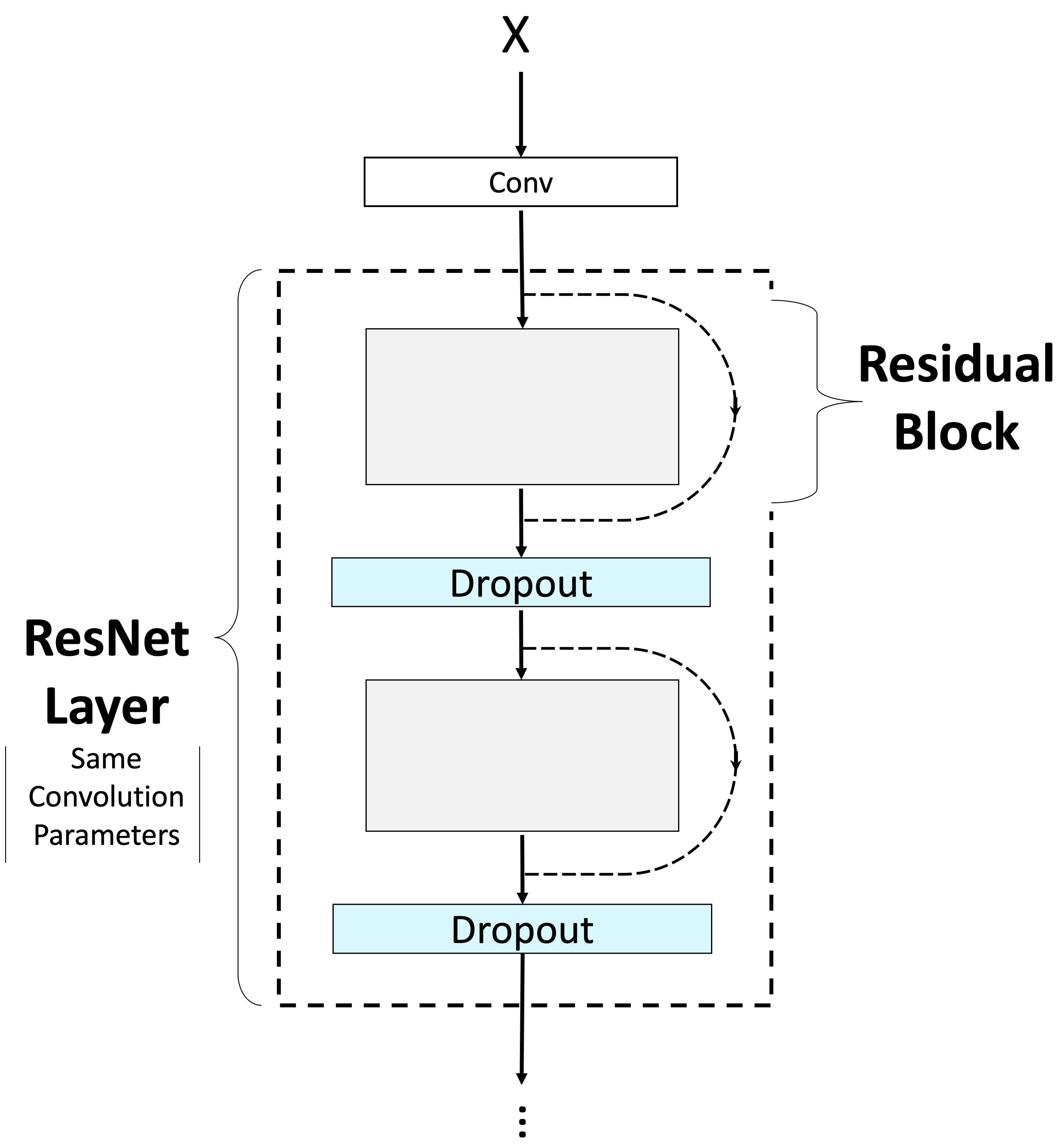}
\vspace{-10pt}
 \caption{A single ResNet layer with dropout layers shown in blue. The ResNet-18 architecture comprises four such layers.}
\label{fig:arc_resnet}
\end{figure}

\paragraph{Classification:} ResNet-18 networks were employed using dataset-specific pre-trained weights and pre-processing protocols from MedMNIST-V2 ~\cite{yang2023medmnist}. The TissueMNIST, PathMNIST, and BloodMNIST datasets were provided as 28x28 voxel inputs. Dropout layers were placed after after each residual block in the pre-trained network (Figure ~\ref{fig:arc_resnet}).
\vspace{-10pt}

\begin{figure}[!htb]
\centering
    \includegraphics[width=0.4\textwidth]{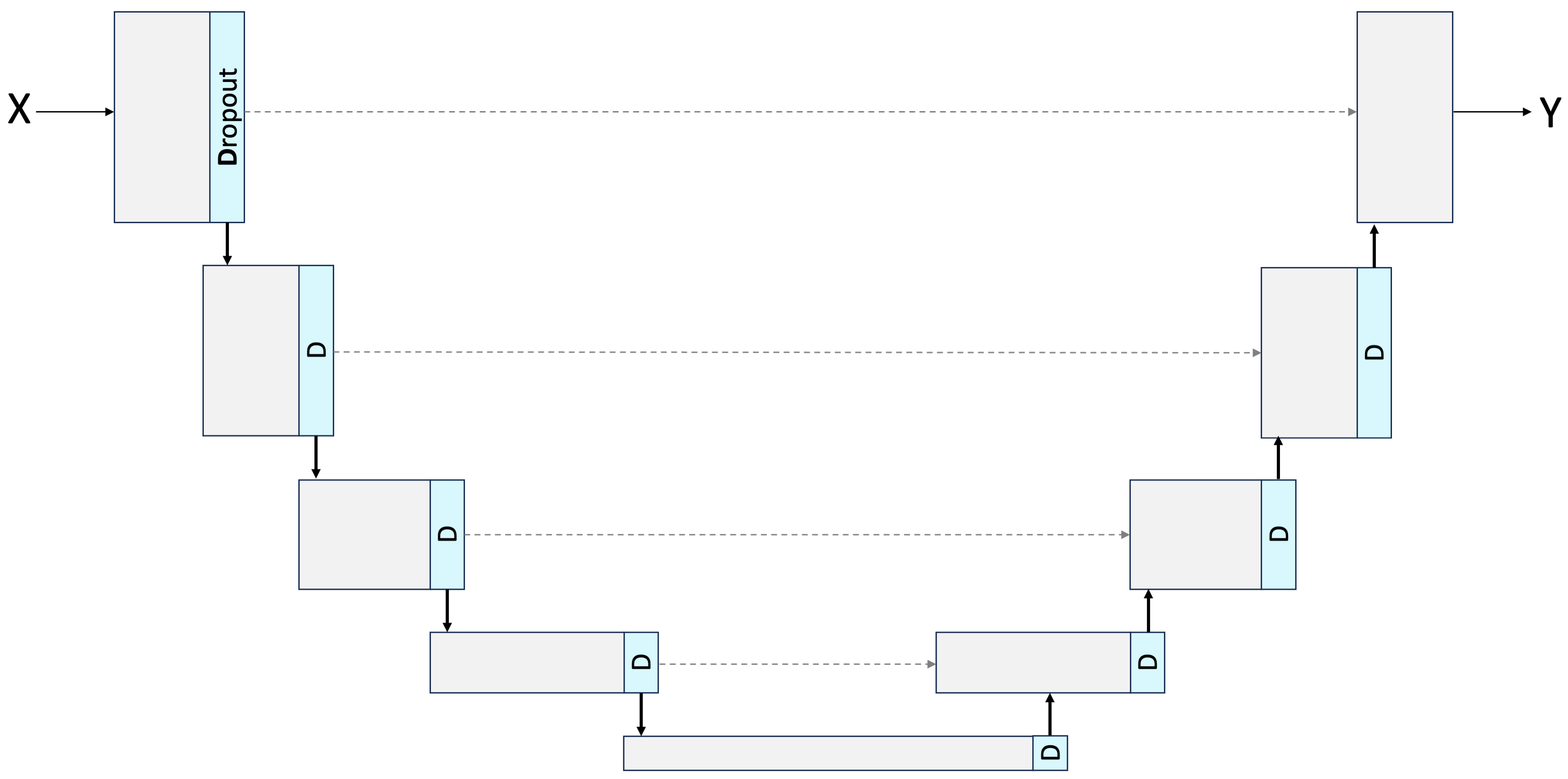}

 \caption{Basic layout of the U-Net network with dropout layers shown in blue.}
\label{fig:arc_unet}
\end{figure}

\paragraph{Segmentation:} The segmentation pipeline employs nnU-Net, using task-specific pre-trained weights from the top-performing models in the Medical Segmentation Decathlon ~\cite{antonelli2022medical}. Each model utilizes a self-configuring segmentation pipeline based on the U-Net architecture with 2D convolutional layers ~\cite{isensee2021nnu}. Dropout layers were placed after each encoding and decoding step in the pre-trained network, except for the output step  (Figure ~\ref{fig:arc_unet}).

\subsubsection{Baseline Dropout Strategies}
\label{sec:suppl_baselines}
In the absence of widely established inference-time dropout strategies, we drew inspiration from training-time dropout approaches, adapting them for inference-time uncertainty estimation. 

\begin{itemize}
    
    \item \textbf{Constant Dropout} ~\cite{srivastava2014dropout}: Dropout rates are fixed across all layers and MC iterations, represented as:
    
    \[
    p^{(l)}_{t} = p \quad \forall \; l, t
    \] 
    
    where \( l \) represents the layer and \( t \) the MC iteration.
    
    \item \textbf{Scheduled (Annealing) Dropout} ~\cite{rennie2014annealed}: Dropout rates are constant across layers but decrease linearly over MC iterations, starting with \( p_{0} = p \) and gradually reducing until dropout is fully disabled. This is given by:
    \[
    p^{(l)}_{t} = p \cdot \left(1 - \frac{t-1}{T-1}\right) \quad \forall \; l, t
    \]
    where \( T \) is the total number of MC iterations.
    
    \item \textbf{Activation-Based Dropout}: Dropout rates vary across layers based on the activation diversity, measured by the coefficient of variation (CoV) of feature map values, while remaining constant across MC iterations. For each layer \( l \) and iteration \( t \), the dropout rate \( p^{(l)}_{t} \) is set as:

    \[
    p^{(l)}_{t} = p \cdot \frac{\text{CoV}(X^{(l)})}   {\max_j \text{CoV}(X^{(j)})}\quad \forall \; l, t 
    \]

    where \( X^{(l)} \) is the feature map for layer \( l \) in the full no-dropout model, and $p$ is the maximum dropout rate applied to the layer with the highest CoV. To handle both positive and negative values in feature maps, CoV is calculated using the mean of absolute values for stability. Our Activation-Based Dropout method is inspired by training-time dropout approaches that adjust dropout rates based on activation patterns during training, as seen in variational dropout ~\cite{kingma2015variational} and adaptive dropout ~\cite{ba2013adaptive}.
\end{itemize}


\subsubsection{Evaluation Metrics}
\label{sec:suppl_eval_metrics}
\paragraph{Predictive Power:} Diluting neural network models can impact their performance. We assessed the deviation from the full (no-dropout) model's prediction performance using the same evaluation metrics as those provided in the original public repository where the dataset was published.

\begin{itemize}

\item \textbf{Mean Squared Error (MSE):} For regression, MSE measures the average squared difference between predicted values and ground truth values, quantifying the error magnitude. The MSE is defined as:

\[
\text{MSE} = \frac{1}{n} \sum_{i=1}^{n} (y_i - \hat{y}_i)^2
\]

where \(y_i\) represents the ground truth value, \(\hat{y}_i\) is the predicted value, and \(n\) is the total number of data points. MSE provides a cohort-level evaluation by averaging the squared errors across all data instances.

In this study, we used the implementation from \texttt{torchmetrics.MeanSquaredError}, with its default parameters.

\item \textbf{Dice Similarity Coefficient (DSC):} For segmentation, DSC measures pixel-wise overlap between predicted binary mask and ground truth regions. The DSC is defined as:

\[
\text{DSC} = \frac{2 |P \cap G|}{|P| + |G|}
\]

where \(P\) is the set of predicted pixels, \(G\) is the set of ground truth pixels, and \(|\cdot|\) denotes the cardinality of the set. DSC was calculated at the pixel level and averaged across all data instances to provide a cohort-level evaluation. 

In this study, we used the implementation from \texttt{torchmetrics.classification.Dice}, with its default parameters.

\item \textbf{Predictive Accuracy (ACC):} For classification, ACC measures the proportion of correctly classified instances over the total number of instances. The ACC is defined as:

\[
\text{ACC} = \frac{\text{Number of Correct Predictions}}{\text{Total Number of Instances}}
\]

In this study, we used the implementation from \texttt{sklearn.metrics.accuracy\_score}.

\end{itemize}

\paragraph{Predictive Uncertainty:}  Uncertainty estimates were assessed as scores that classify model predictions into two outcomes: reject or do-not-reject, following ~\cite{mukhoti2018evaluating, krishnan2020improving, mobiny2021dropconnect}. To evaluate the alignment of uncertainty estimates with model prediction errors, we used:

\begin{itemize}
\item \textbf{Expected Calibration Error (ECE)}: For segmentation uncertainty, ECE measures calibration by aligning uncertainty scores with actual model errors ~\cite{laves2019uncertainty}. The ECE is defined as:

\[
\text{ECE} = \sum_{m=1}^M \frac{|B_m|}{N} \left| \text{uncert}(B_m) - \text{err}(B_m) \right|
\]

where \(M\) is the total number of bins, \(B_m\) is the set of pixels in bin \(m\), \(|B_m|\) is the number of pixels in \(B_m\), \(N\) is the total number of pixels, \(\text{uncert}(B_m)\) is the average predicted uncertainty score, and \(\text{err}(B_m)\) is the actual model error within the bin. ECE was calculated at the pixel level and averaged across all data to provide a cohort-level evaluation.

In this study, we used the implementation from \texttt{torchmetrics.CalibrationError} with parameters \texttt{CalibrationError(task='binary', norm='l1', n\_bins=15)} and all other parameters set to their default values.

\item \textbf{Area Under the Accuracy-Rejection Curve (AUARC)}: For classification uncertainty, AUARC evaluates the trade-off between predictive accuracy and the fraction of predictions rejected based on uncertainty scores ~\cite{geifman2018bias, nadeem2009accuracy}. The AUARC is defined as:

\[
\text{AUARC} = \int_0^1 \text{ACC}(r) \, dr
\]

where \(r\) is the rejection fraction, and \(\text{ACC}(r)\) is the accuracy of the retained predictions at rejection fraction \(r\). Uncertainty scores corresponded to the highest predicted class was used. Accuracy was calculated for rejection thresholds corresponding to each uncertainty score percentile, and the area under the curve was computed using \texttt{sklearn.metrics.auc}. Higher AUARC values indicate that the model maintains high accuracy while effectively rejecting uncertain predictions.

\item \textbf{Boundary Uncertainty Consistency (BUC):} Accurate estimation of uncertainty at organ boundaries is critical in medical imaging, as these regions often exhibit high uncertainty due to anatomical variability, image noise, and ambiguous structures~\cite{yue2024boundary, mehrtash2020confidence,li2024boundary}.
BUC quantifies the proportion of total uncertainty concentrated along boundaries compared to interior regions.

\[
\text{BUC} = \frac{\text{Mean}(\text{uncert}_{\text{Boundary}})}{\text{Mean}(\text{uncert}_{\text{Boundary}}) + \text{Mean}(\text{uncert}_{\text{Interior}})},
\]

where \(\text{Mean}(\text{uncert}_{\text{Boundary}})\) and \(\text{Mean}(\text{uncert}_{\text{Interior}})\) represent the average uncertainties in boundary and interior regions, respectively. The boundary region in our experiments was defined as a 5-pixel-wide band surrounding the edges of the segmented objects.

BUC highlights the model's ability to assign higher uncertainty to critical boundary areas, ensuring robust uncertainty evaluation essential for clinical decision-making.

\item \textbf{Interval Efficiency Ratio (IER):} Measures the trade-off between predictive interval width and its coverage probability (PICP), quantifying the efficiency of uncertainty estimation. The interval width represents the average range of the prediction intervals and is calculated as:

\[
\text{Uncertainty Interval Width} = \frac{1}{N} \sum_{i=1}^N 2 \cdot Z \cdot \sigma_i
\]

where \( Z \) represents the Z-score corresponding to the desired confidence level (e.g., \( Z = 1.96 \) for 95\% confidence intervals), and \( \sigma_i \) denotes the predicted standard deviation, calculated as the standard deviation of the Monte Carlo (MC) predictions at each data point \( i \).

The PICP quantifies the proportion of true values \( y_{\text{true}} \) that fall within the predicted intervals and is defined as:

\[
\text{PICP} = \frac{1}{N} \sum_{i=1}^N \mathbb{I} \left[ y_{\text{true}, i} \in \left( \mu_i - Z \cdot \sigma_i, \mu_i + Z \cdot \sigma_i \right) \right]
\]

where \( \mathbb{I} \) is the indicator function, \( \mu_i \) is the predicted mean, and \( N \) is the number of samples.

The ratio of interval width to PICP evaluates the trade-off between the tightness of prediction intervals and the coverage of true values, with lower ratios indicating more efficient uncertainty estimation.

\end{itemize}

\section{Appendix B: Additional Results}
\label{suppl:appendix_B}

\subsection{Synthetic Data}
\label{supp:synthetic_data}
\begin{figure*}[t]
    \includegraphics[width=\linewidth]{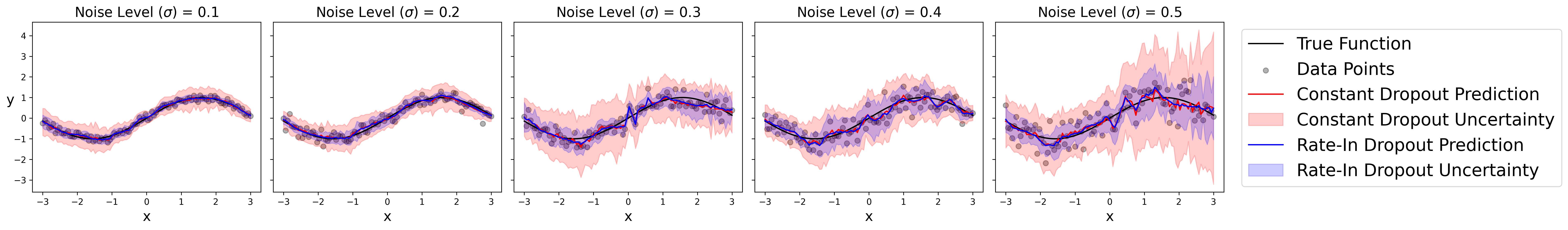}
    \vspace{-15pt}
    \caption{\small Rate-In dropout yields narrower, more stable uncertainty intervals under increasing noise levels ($\sigma = 0.1-0.5$), compared to constant dropout. Black: true function and training data; Red: constant dropout ($p=0.10$); Blue: Rate-In dropout ($\epsilon=0.10$).}

    \label{fig:suppl_scatter_noise}
\end{figure*}

\begin{figure}[H]
    \centering
    \begin{subfigure}{0.32\columnwidth}
        \includegraphics[width=\linewidth]{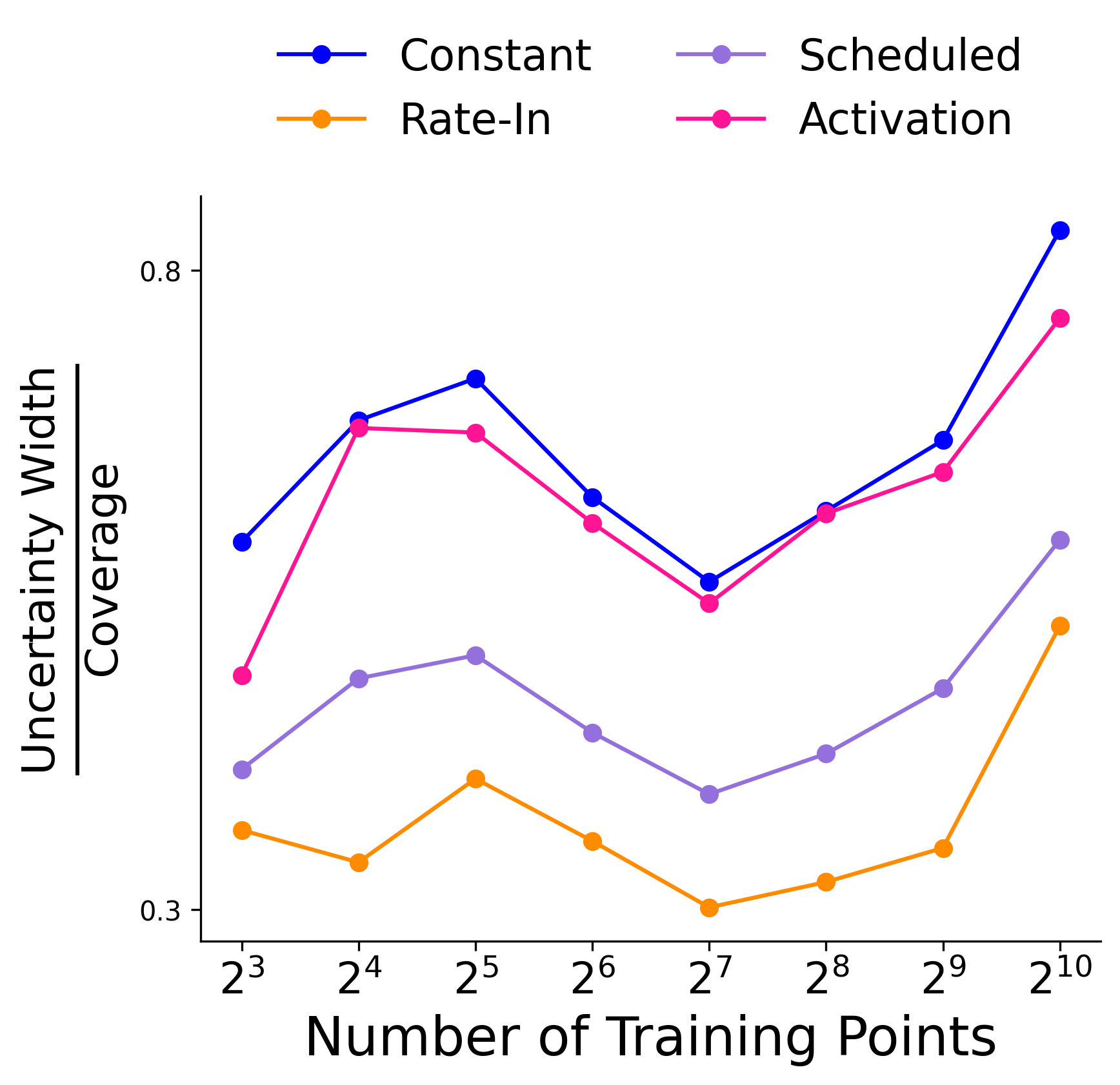}
        \caption{\scriptsize $\sigma = 0.01$}
        \label{fig:sub1}
    \end{subfigure}
    \hfill
    \begin{subfigure}{0.32\columnwidth}
        \includegraphics[width=\linewidth]{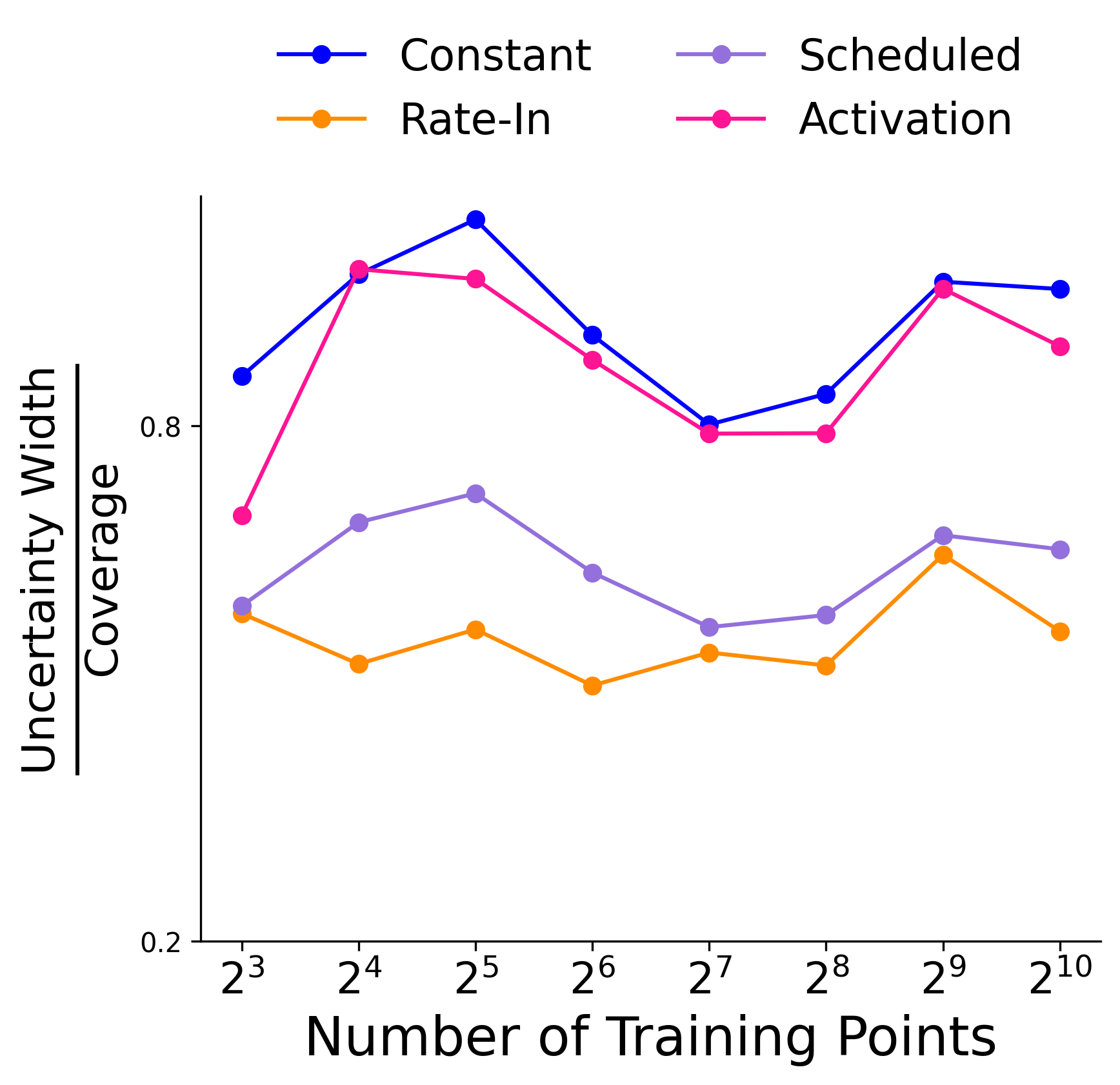}
        \caption{\scriptsize $\sigma = 0.10$}
        \label{fig:sub2_}
    \end{subfigure}
    \hfill
    \begin{subfigure}{0.32\columnwidth}
        \includegraphics[width=\linewidth]{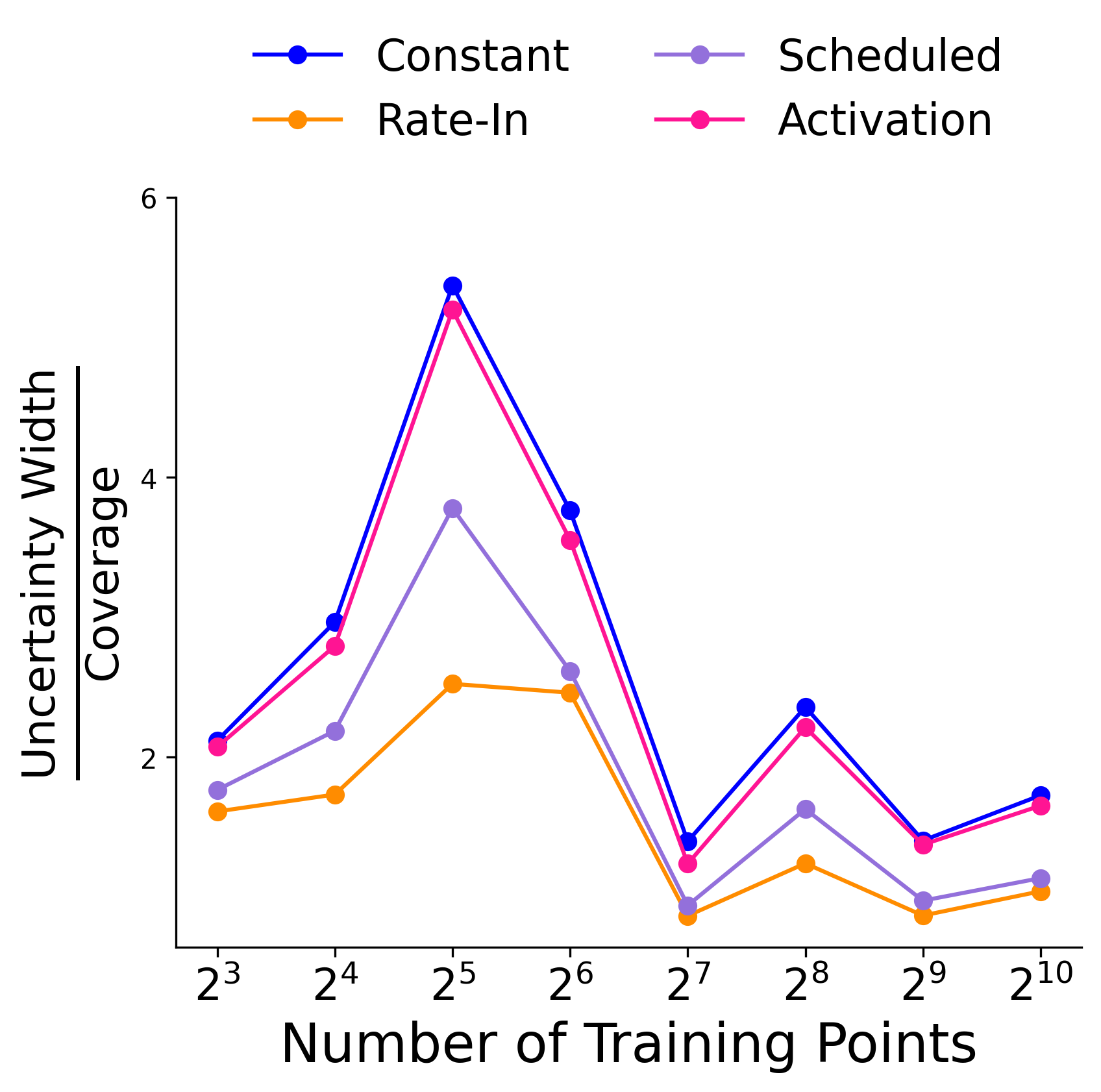}
        \caption{\scriptsize $\sigma = 0.50$}
        \label{fig:sub3}
    \end{subfigure}

    \caption{\small \textbf{Rate-In is more efficient in uncertainty estimation.} The ratio of uncertainty interval width to 95\% coverage for varying number of training points at fixed noise levels. Lower ratios indicate more efficient uncertainty estimation. (a) $\sigma = 0.01$, (b) $\sigma = 0.10$, and (c) $\sigma = 0.50$.}
    \label{fig:suppl_int_eff}
\end{figure}

Figures~\ref{fig:suppl_scatter_noise} and ~\ref{fig:suppl_int_eff} evaluate Rate-In dropout's performance. Figure~\ref{fig:suppl_scatter_noise} compares uncertainty intervals from constant dropout ($p = 0.10$, red) and Rate-In dropout ($\epsilon = 0.10$, blue) across five noise levels ($\sigma = 0.1-0.5$), showing Rate-In's more precise and stable bounds under increasing noise. Figure~\ref{fig:suppl_int_eff} examine performance at fixed noise levels ($\sigma = 0.01, 0.10, 0.50$) across varying training set sizes, showing uncertainty interval efficiency ratios.

\begin{figure}[!htb]
\centering
    \includegraphics[width=1.0\columnwidth]{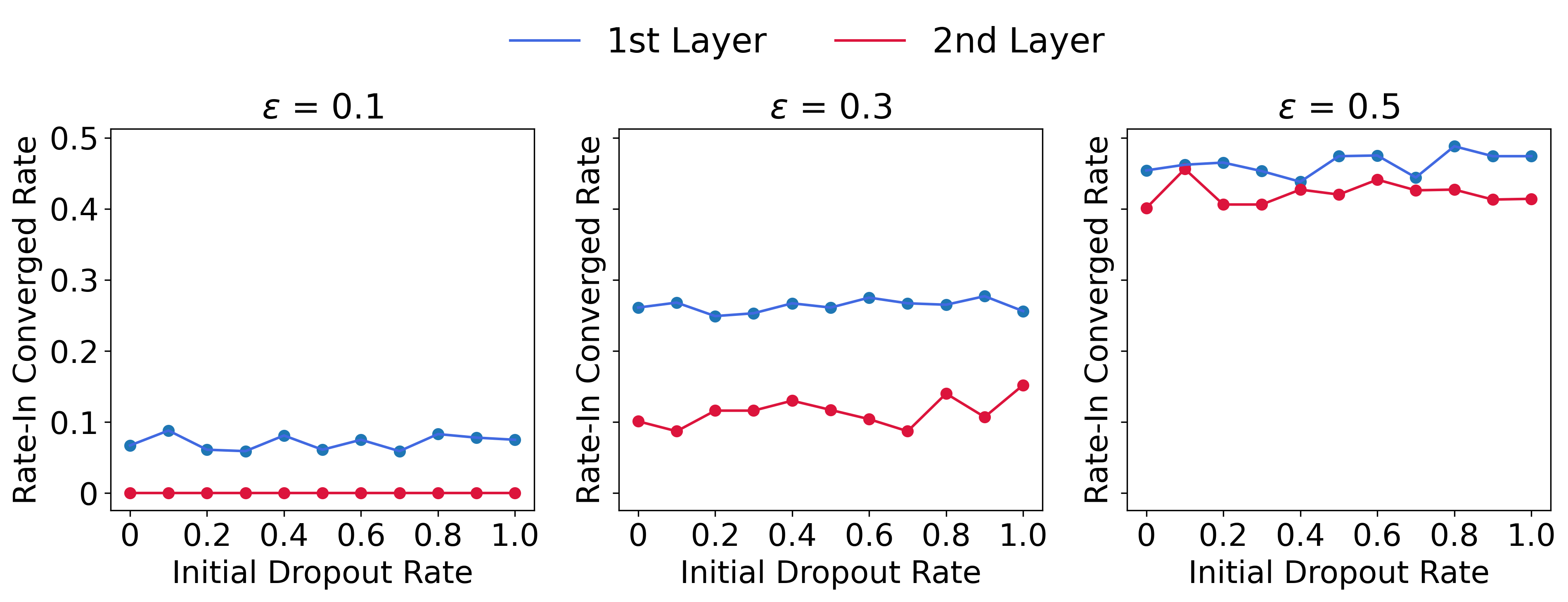}

\caption{\small Rate-In final dropout rate ($y$-axis) versus initial dropout rate ($x$-axis) for different information loss thresholds ($\epsilon$ = 0.10, 0.30, 0.50). Parameters: $N_{max}$ = 100, $\sigma$ = 0.50.}
\label{fig:suppl_initial_rates}
\end{figure}

\noindent Figure ~\ref{fig:suppl_initial_rates} analyzes the convergence behavior of Rate-In across varying initial dropout rates under different information loss thresholds ($\epsilon$ = 0.10, 0.30, 0.50), with $N_{max}$ = 100 and $\sigma$ = 0.50. The results demonstrate that Rate-In converges to consistent dropout rates, independent of the initial dropout rate values.

\subsection{Classification}
\begin{figure}[ht]
    \centering
    \includegraphics[width=1.0\columnwidth]{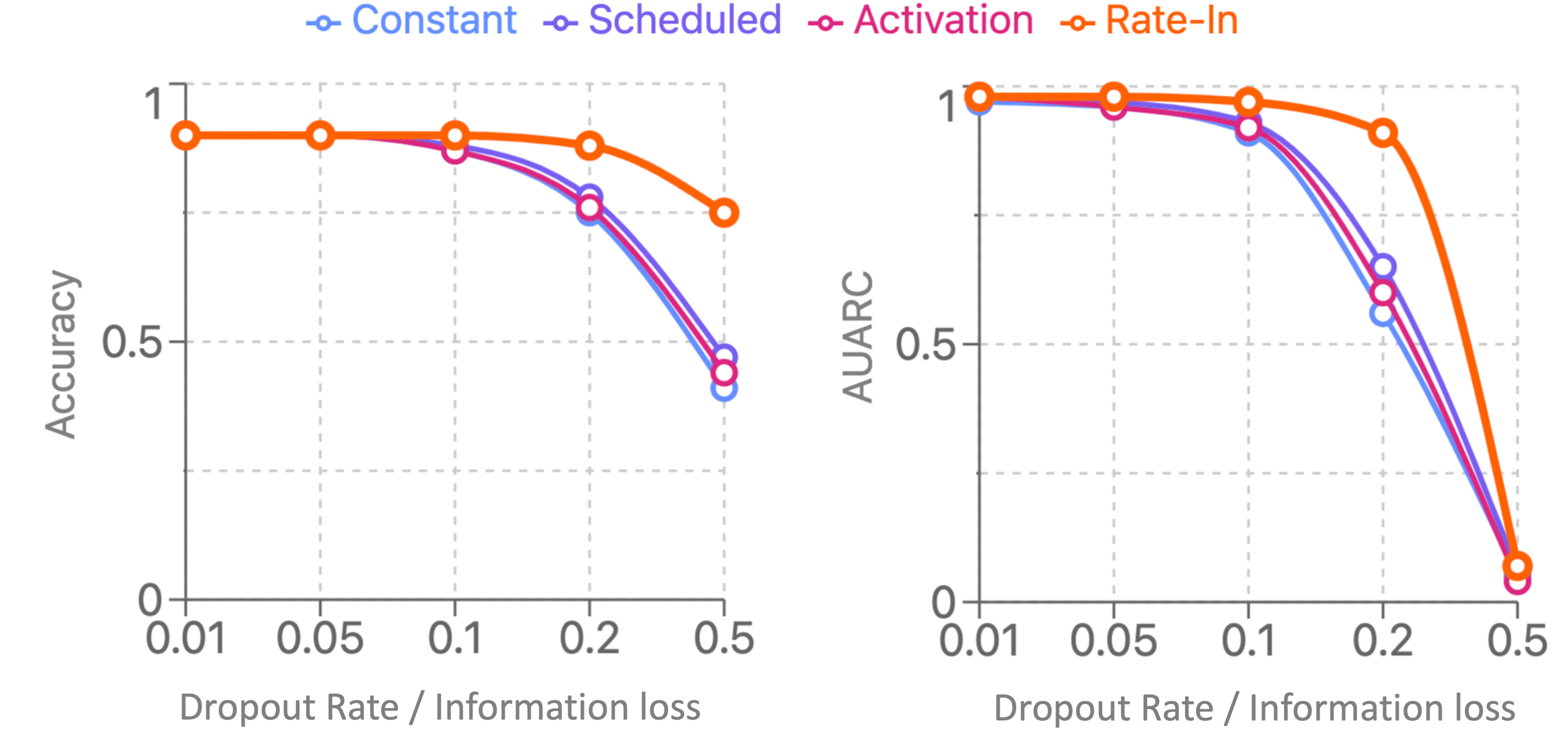}
   \caption{\small \textbf{Rate-In maintains high classification accuracy even at aggressive dropout rates where conventional methods significantly degrade.} 
    Classification accuracy on PathMNIST dataset comparing different dropout strategies across increasing dropout rates. 
    While traditional methods show sharp accuracy decline, Rate-In preserves performance by adaptively adjusting dropout patterns based on local feature importance, demonstrating its robustness for reliable medical image classification.}
 
    \label{fig:PathMNIST}
\end{figure}

Figure~\ref{fig:PathMNIST} presents the performance of Rate-In on the PathMNIST dataset, demonstrating its ability to maintain high accuracy and better uncertainty calibration compared to baseline methods. Notably, Rate-In sustains strong performance across varying levels of information loss.

\subsection{Segmentation}
Table ~\ref{tab:suppl_segmentation} presents the comparison between Rate-In and all baseline dropout approaches across segmentation tasks. 

\begin{table*}[t]
\caption{Comparison of Rate-In and baseline dropout approaches for uncertainty estimation and segmentation accuracy across anatomical zones in MRI and CT modalities, evaluated using DSC, ECE, and BUC metrics at varying dropout rates. $^*$ Best baseline dropout method is defined by the highest average DSC across dropout rates per task. The overall \textbf{best-performing approach across metrics is in bold}} \vspace{-10pt}
\label{tab:suppl_segmentation}
\centering
\footnotesize
\setlength{\tabcolsep}{3.0pt}
\begin{tabular}{@{}lllcccccccccc@{}}
\toprule
\multirow{2}{*}{Zone} & \multirow{2}{*}{Modality} & \multirow{2}{*}{Dropout Method} & DSC & \multicolumn{3}{c}{DSC (Full Model, \% change)} & \multicolumn{3}{c}{ECE $\times 10^{-3}$} & \multicolumn{3}{c}{BUC} \\
\cmidrule(lr){5-7} \cmidrule(lr){8-10} \cmidrule(lr){11-13}
& & & Full Model & $p\text{,}\epsilon{\text{=}}0.01$ & $p\text{,}\epsilon{\text{=}}0.05$ & $p\text{,}\epsilon{\text{=}}0.10$ & $p\text{,}\epsilon{\text{=}}0.01$ & $p\text{,}\epsilon{\text{=}}0.05$ & $p\text{,}\epsilon{\text{=}}0.10$ & $p\text{,}\epsilon{\text{=}}0.01$ & $p\text{,}\epsilon{\text{=}}0.05$ & $p\text{,}\epsilon{\text{=}}0.10$ \\
\midrule
\multirow{2}{*}{Peripheral} & \multirow{2}{*}{MRI} & \textbf{Rate-In} & \multirow{4}{*}{0.682} & +0.29\% & +2.79\% & +2.35\% & 4.60 & \textbf{4.26} & \textbf{4.58} & \textbf{0.67} & \textbf{0.63} & \textbf{0.61} \\
& & Constant$^*$ & & +2.05\% & +2.59\% & +2.05\% &\textbf{4.40} & 5.547 & 7.85 & 0.54 & 0.48 & 0.45 \\
& & Scheduled & & +1.06\% & +2.84\% & +2.51\% & 4.44 & 5.14 & 7.24 & 0.55 & 0.47 & 0.45 \\
& & Activation & & +1.17\% & +2.79\% & +2.64\% & 4.62 & 4.40 & 5.01 & 0.58 & 0.53 & 0.50 \\
\midrule
\multirow{2}{*}{Transitional} & \multirow{2}{*}{MRI} & \textbf{Rate-In} & \multirow{4}{*}{0.892} & -0.34\% & -0.78\% & -0.67\% & \textbf{3.97} & \textbf{4.03} & \textbf{4.86} & \textbf{0.87} & \textbf{0.84} & \textbf{0.80} \\
& & Constant$^*$ & & +0.11\% & -0.78\% & -0.78\% & 4.25 & 6.60 & 10.73 & 0.70 & 0.62 & 0.57 \\
& & Scheduled & & -0.16\% & -0.69\% & -0.74\% & 4.30 & 6.15 & 9.64 & 0.71 & 0.63 & 0.58 \\
& & Activation & & -0.18\% & -0.71\% & -0.69\% & 4.13 & 4.55 & 5.76 & 0.78 & 0.71 & 0.67 \\
\midrule
\multirow{2}{*}{Liver} & \multirow{2}{*}{CT} & \textbf{Rate-In} & \multirow{4}{*}{0.955} & +1.05\% & +1.05\% & +0.94\% & \textbf{6.73} & \textbf{5.54} & \textbf{4.65} & \textbf{0.93} & \textbf{0.93} & \textbf{0.92} \\
& & Constant & & -0.00\% & -0.01\% & -0.03\% & 6.80 & 5.80 & 4.90 & 0.92 & 0.91 & 0.91 \\
& & Scheduled$^*$ & & +0.00\% & -0.00\% & -0.00\% & 6.90 & 5.80 & 5.00 & 0.92 & 0.91 & 0.91 \\
& & Activation & & -0.00\% & -0.03\% & -0.04\% & 7.00 & 6.20 & 5.40 & 0.92 & 0.91 & 0.91 \\
\midrule
\multirow{2}{*}{Tumor} & \multirow{2}{*}{CT} & \textbf{Rate-In} & \multirow{4}{*}{0.579} & +1.21\% & -0.52\% & -2.59\% & \textbf{1.78} & \textbf{1.78} & \textbf{1.88} & \textbf{0.61} & 0.53 & 0.48 \\
& & Constant & & -0.12\% & -2.14\% & -2.61\% & 1.80 & 1.80 & 1.90 & 0.59 & 0.52 & 0.49 \\
& & Scheduled & & +0.05\% & -1.96\% & -2.62\% & 1.80 & 1.80 & 2.00 & 0.59 & 0.52 & 0.49 \\
& & Activation$^*$ & & +0.17\% & -1.90\% & -1.73\% & 1.80 & 1.80 & 1.90 & 0.60 & \textbf{0.55} & \textbf{0.53} \\
\bottomrule

\end{tabular}

\end{table*}

\subsection{Out-of-Distribution (OOD) Tasks}

Table~\ref{tab:ood_supp} reports the mean accuracy (AUARC) across all corruption types in the MedMNIST-C datasets for Rate-In and all benchmark methods.

\begin{table*}[htbp]
\scriptsize  
\centering
\renewcommand{\arraystretch}{0.9}

\caption{\footnotesize Rate-In robustness evaluation across MedMNIST-C corruption types. The table compares Rate-In with benchmark methods across three datasets, showing superior performance at varying dropout rates ($p$,$\epsilon$=0.05-0.20), with average ACC(AUARC) highlighted in the 'Avg.' column.}\vspace{-10pt}

\setlength{\tabcolsep}{6.0pt}
\begin{tabular}{@{}lcl|c|cccccccccc@{}}
\toprule
\textbf{Dataset} & \textbf{$p,\epsilon$} & \textbf{Method} & \textbf{Avg.} & \textbf{bright.} & \textbf{bright.} & \textbf{bubble} & \textbf{contrast} & \textbf{contrast} & \textbf{defocus} & \textbf{jpeg} & \textbf{motion} & \textbf{pixelate} & \textbf{saturate} \\
& & & & \textbf{down} & \textbf{up} & & \textbf{down} & \textbf{up} & \textbf{blur} & \textbf{compr.} & \textbf{blur} & & \\
\midrule

\multirow{14}{*}{\textit{BloodMNIST}} 
& \multicolumn{2}{l|}{ACC Full Model} & .77 & .87 & .88 & .18 & .88 & .92 & .69 & .87 & .70 & .91 & .83 \\
\cmidrule(lr){2-14}
& \multirow{4}{*}{$0.05$} & Rate-In & \textbf{.77}(.87) & .86(.96) & \textbf{.84}(.94) & .19(.17) & .91(.97) & \textbf{.91}(.98) & .70(.89) & \textbf{.90}(.97) & \textbf{.70}(.88) & \textbf{.92}(.98) & \textbf{.81}(.95) \\
& & Constant & .76(.87) & .86(.97) & .71(.88) & .19(.20) & \textbf{.94}(.98) & .87(.98) & \textbf{.74}(.89) & .87(.97) & .69(.89) & .90(.98) & .79(.93) \\
& & Scheduled & .76(.87) & .86(.97) & .77(.90) & .19(.19) & \textbf{.94}(.98) & .89(.98) & .72(.88) & .88(.97) & .68(.88) & .90(.98) & .80(.93) \\
& & Activation & \textbf{.77}(.87) & .86(.97) & .77(.90) & .19(.20) & \textbf{.94}(.98) & .89(.98) & .72(.89) & .88(.97) & \textbf{.70}(.88) & \textbf{.92}(.98) & \textbf{.81}(.94) \\
\cmidrule(lr){2-14}

& \multirow{4}{*}{$0.10$} & Rate-In & \textbf{.75}(.86) & \textbf{.84}(.96) & \textbf{.72}(.89) & \textbf{.20}(.18) & .94(.98) & \textbf{.86}(.98) & .72(.88) & \textbf{.87}(.97) & .67(.88) & \textbf{.90}(.98) & \textbf{.78}(.93) \\
& & Constant & .70(.83) & .80(.95) & .45(.67) & .18(.21) & \textbf{.96}(.98) & .82(.95) & \textbf{.73}(.86) & .85(.97) & .71(.88) & .81(.96) & .73(.89) \\
& & Scheduled & .73(.84) & .83(.96) & .56(.72) & .19(.22) & \textbf{.96}(.99) & .84(.96) & \textbf{.73}(.87) & .85(.97) & \textbf{.72}(.88) & .86(.97) & .76(.89) \\
& & Activation & .72(.85) & .81(.96) & .55(.74) & .18(.21) & .95(.99) & \textbf{.86}(.97) & .72(.87) & .85(.97) & .69(.88) & .87(.97) & .76(.91) \\
\cmidrule(lr){2-14}

& \multirow{4}{*}{$0.20$} & Rate-In & \textbf{.66}(.79) & \textbf{.73}(.89) & \textbf{.35}(.52) & \textbf{.17}(.20) & \textbf{.95}(.98) & \textbf{.72}(.87) & \textbf{.73}(.86) & \textbf{.76}(.92) & \textbf{.68}(.85) & \textbf{.78}(.94) & \textbf{.71}(.86) \\
& & Constant & .45(.55) & .63(.80) & .17(.23) & .12(.17) & .72(.89) & .28(.37) & .64(.71) & .38(.49) & .58(.66) & .54(.64) & .46(.58) \\
& & Scheduled & .56(.66) & .72(.87) & .19(.29) & .12(.17) & .87(.94) & .48(.58) & .66(.77) & .65(.74) & .63(.77) & .70(.80) & .61(.70) \\
& & Activation & .59(.73) & .72(.88) & .19(.28) & .12(.17) & .87(.96) & .62(.79) & .67(.84) & .69(.85) & .67(.83) & .74(.90) & .66(.80) \\
\midrule

\multirow{14}{*}{\textit{PathMNIST}}
& \multicolumn{2}{l|}{ACC Full Model} & .54 & .44 & .42 & .67 & .51 & .79 & .34 & .25 & .53 & .78 & .71 \\
\cmidrule(lr){2-14}
& \multirow{4}{*}{$0.05$} & Rate-In & \textbf{.52}(.62) & \textbf{.43}(.38) & \textbf{.39}(.53) & \textbf{.63}(.81) & .53(.61) & \textbf{.75}(.82) & \textbf{.33}(.37) & \textbf{.24}(.38) & \textbf{.52}(.63) & \textbf{.75}(.91) & \textbf{.63}(.78) \\
& & Constant & .47(.58) & .42(.35) & .34(.48) & .53(.71) & \textbf{.54}(.61) & .69(.78) & .23(.35) & .21(.35) & .50(.60) & .70(.85) & .59(.70) \\
& & Scheduled & .49(.59) & \textbf{.43}(.35) & .35(.48) & .58(.73) & \textbf{.54}(.60) & .72(.78) & .24(.35) & .22(.36) & .51(.62) & .72(.87) & \textbf{.63}(.73) \\
& & Activation & .49(.58) & .42(.34) & .35(.48) & .56(.72) & \textbf{.54}(.61) & .70(.78) & .23(.36) & .21(.36) & \textbf{.52}(.60) & .71(.86) & .62(.72) \\
\cmidrule(lr){2-14}

& \multirow{4}{*}{$0.10$} & Rate-In & \textbf{.50}(.59) & \textbf{.41}(.35) & \textbf{.35}(.51) & \textbf{.59}(.76) & \textbf{.55}(.59) & \textbf{.71}(.79) & \textbf{.29}(.37) & \textbf{.22}(.36) & \textbf{.52}(.60) & \textbf{.72}(.88) & \textbf{.63}(.74) \\
& & Constant & .37(.47) & .36(.27) & .27(.36) & .38(.47) & .49(.60) & .54(.60) & .16(.28) & .20(.32) & .38(.50) & .55(.70) & .38(.55) \\
& & Scheduled & .39(.49) & .39(.28) & .28(.39) & .41(.52) & .49(.62) & .58(.65) & .18(.30) & .19(.33) & .39(.53) & .58(.74) & .46(.57) \\
& & Activation & .38(.48) & .38(.28) & .28(.38) & .41(.49) & .49(.60) & .57(.63) & .17(.31) & .19(.34) & .39(.53) & .55(.72) & .40(.56) \\
\cmidrule(lr){2-14}

& \multirow{4}{*}{$0.20$} & Rate-In & \textbf{.39}(.49) & \textbf{.31}(.26) & \textbf{.29}(.41) & \textbf{.40}(.54) & \textbf{.49}(.58) & \textbf{.52}(.61) & \textbf{.24}(.35) & .16(.29) & \textbf{.50}(.58) & \textbf{.54}(.72) & \textbf{.42}(.54) \\
& & Constant & .21(.26) & .13(.17) & .16(.14) & .23(.18) & .30(.49) & .25(.23) & .16(.23) & .19(.25) & .19(.28) & .32(.29) & .18(.29) \\
& & Scheduled & .25(.31) & .20(.19) & .17(.19) & \textbf{.24}(.24) & .38(.55) & .31(.33) & .16(.25) & \textbf{.22}(.31) & .25(.34) & \textbf{.33}(.33) & \textbf{.26}(.33) \\
& & Activation & .24(.29) & .19(.17) & .19(.18) & .22(.22) & .35(.52) & .29(.26) & .17(.25) & .21(.33) & .25(.34) & .32(.29) & .21(.31) \\
\midrule

\multirow{14}{*}{\textit{TissueMNIST}}
& \multicolumn{2}{l|}{ACC Full Model} & .65 & .65 & .72 & --- & .70 & .70 & --- & .55 & --- & .56 & --- \\
\cmidrule(lr){2-14}
& \multirow{4}{*}{$0.05$} & Rate-In & \textbf{.63}(.77) & \textbf{.60}(.78) & \textbf{.70}(.85) & --- & \textbf{.72}(.86) & \textbf{.64}(.77) & --- & \textbf{.58}(.61) & --- & \textbf{.56}(.74) & --- \\
& & Constant & .50(.54) & .45(.42) & .57(.70) & --- & .59(.73) & .47(.50) & --- & .42(.42) & --- & .48(.49) & --- \\
& & Scheduled & .51(.60) & .46(.51) & .59(.75) & --- & .59(.73) & .49(.54) & --- & .43(.48) & --- & .51(.56) & --- \\
& & Activation & .50(.56) & .46(.46) & .57(.72) & --- & .60(.74) & .46(.49) & --- & .43(.43) & --- & .49(.52) & --- \\
\cmidrule(lr){2-14}

& \multirow{4}{*}{$0.10$} & Rate-In & \textbf{.59}(.71) & \textbf{.58}(.71) & \textbf{.66}(.80) & --- & \textbf{.67}(.84) & \textbf{.58}(.69) & --- & \textbf{.48}(.52) & --- & \textbf{.56}(.69) & --- \\
& & Constant & .30(.27) & .29(.21) & .41(.44) & --- & .36(.33) & .28(.23) & --- & .22(.19) & --- & .24(.25) & --- \\
& & Scheduled & .34(.35) & .30(.25) & .43(.49) & --- & .42(.45) & .32(.33) & --- & .28(.32) & --- & .29(.29) & --- \\
& & Activation & .30(.29) & .29(.22) & .42(.43) & --- & .35(.35) & .30(.24) & --- & .23(.23) & --- & .24(.30) & --- \\
\cmidrule(lr){2-14}

& \multirow{4}{*}{$0.20$} & Rate-In & \textbf{.45}(.48) & \textbf{.46}(.47) & \textbf{.45}(.50) & --- & \textbf{.56}(.68) & \textbf{.37}(.39) & --- & \textbf{.41}(.34) & --- & \textbf{.43}(.51) & --- \\
& & Constant & .11(.06) & .11(.08) & .11(.05) & --- & .12(.06) & .10(.08) & --- & .13(.08) & --- & .08(.03) & --- \\
& & Scheduled & .12(.07) & .12(.09) & .15(.10) & --- & .13(.07) & .11(.09) & --- & .13(.07) & --- & .08(.02) & --- \\
& & Activation & .11(.07) & .10(.08) & .12(.06) & --- & .13(.07) & .11(.09) & --- & .13(.08) & --- & .08(.02) & --- \\

\bottomrule
\end{tabular}
\label{tab:ood_supp}
\end{table*}

\subsubsection{Natural Scene Images}
We employed the ImageNet dataset from the ImageNet Object Localization Challenge \cite{howard2018imagenet}, which consists of 50,000 natural scene images spanning 1,000 categories. For our experiments, we used the PyTorch implementations of MobileNet-V3-Large, VGG-16, and ResNet-50, along with their pre-trained IMAGENET1K\_V1 weights \cite{paszke2019pytorch}. Dropout layers were introduced after each convolutional layer in all three networks.

\begin{figure}[H]

\centering
    \includegraphics[width=0.48\textwidth]{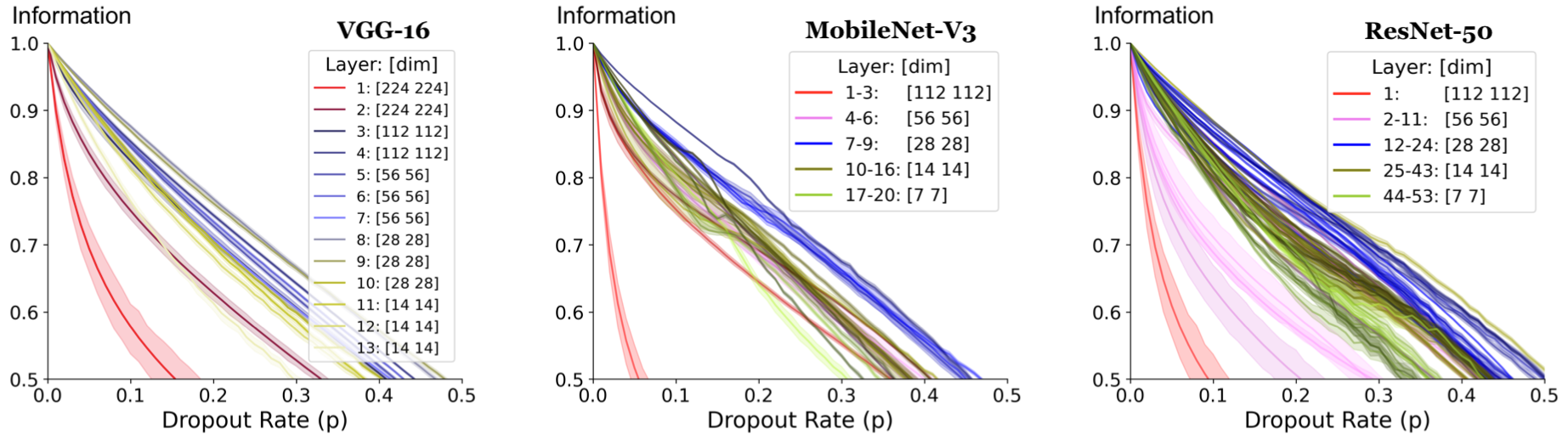}

 \caption{\small Information loss patterns across VGG-16, MobileNet, and ResNet architectures using ImageNet data.}
 
\label{fig:image-net}
\vspace{-10pt}
\end{figure}

Figure \ref{fig:image-net} illustrates the information loss patterns induced by dropout across all layers of the three neural network architectures. Dropout led to substantial information loss in early layers, with a 50\% reduction for rates between 0.05 and 0.15. ResNet-50 showed the greatest variability across layers.

\subsection{Computational Complexity Analysis} 
\label{supp:complexity_analysis}
The Rate-In algorithm may introduce an additional computational overhead during inference, resulting in increased latency from input availability to prediction. Section~\ref{sec:implementation_considerations} outlines strategies to mitigate this inference-time overhead.  

The pseudocode in Listing~\ref{lst:adaptive_dropout} illustrates the \texttt{AdaptiveInformationDropout} class, which implements the optimization process for the Rate-In algorithm. The computational complexity of the dropout rate optimization is \(O(N_{\text{max}} \cdot (n + f))\), where \(N_{\text{max}}\) is the maximum number of optimization iterations, \(n\) is the number of input instances, and \(f\) is the complexity of the information loss calculation, which may depend on \(n\).  

The optimization process dominates the complexity, iterating \(N_{\text{max}}\) times with \(O(n)\) dropout operations and \(O(f)\) calculations per iteration. The computational complexity of each dropout operation, while initially appearing as \(O(n)\), further breaks down to \(O(d)\), where \(d\) is the number of elements in each input instance. Thus, the true complexity of a single dropout operation is \(O(n \cdot d)\).  

Once the dropout rates are determined, they remain fixed during Monte Carlo inference and, therefore, do not introduce additional complexity compared to regular constant dropout approaches.  

\begin{lstlisting} [caption={AdaptiveInformationDropout Class implementation pseudocode for the Rate-In algorithm}, label={lst:adaptive_dropout}]
class AdaptiveInformationDropout(torch.nn.Module):
    def __init__(self, ...):
        # O(1)
        ...

    def _apply_dropout(self, x: torch.Tensor, rate: float) -> torch.Tensor:
        # O(n), n = number of elements in x
        return torch.nn.functional.dropout(x, p=rate, training=self.training)

    def _optimize_dropout_rate(self, x: torch.Tensor) -> float:
        # O(N_max * (n + f))
        # N_max = max_iterations, n = elements in x, f = complexity of calc_information_loss
        for iteration in range(config.max_iterations):  # O(N_max)
            post_dropout = self._apply_dropout(pre_dropout, current_rate)  # O(n)
            info_loss = self.calc_information_loss(...)  # O(f)
            # Update dropout rate: O(1)
            ...

    def forward(self, x: torch.Tensor) -> torch.Tensor:
        if self.training:
            # O(N_max * (n + f) + n) = O(N_max * (n + f))
            optimized_rate = self._optimize_dropout_rate(x)  # O(m * (n + f))
            return self._apply_dropout(x, optimized_rate)  # O(n)
        return x  # O(1) if not training

# Overall forward pass complexity: O(N_max * (n + f))

\end{lstlisting}

\begin{figure}[H]
    \centering
    \begin{minipage}{0.95\columnwidth}
        \centering
        \begin{subfigure}{0.45\columnwidth}
            \includegraphics[width=\linewidth]{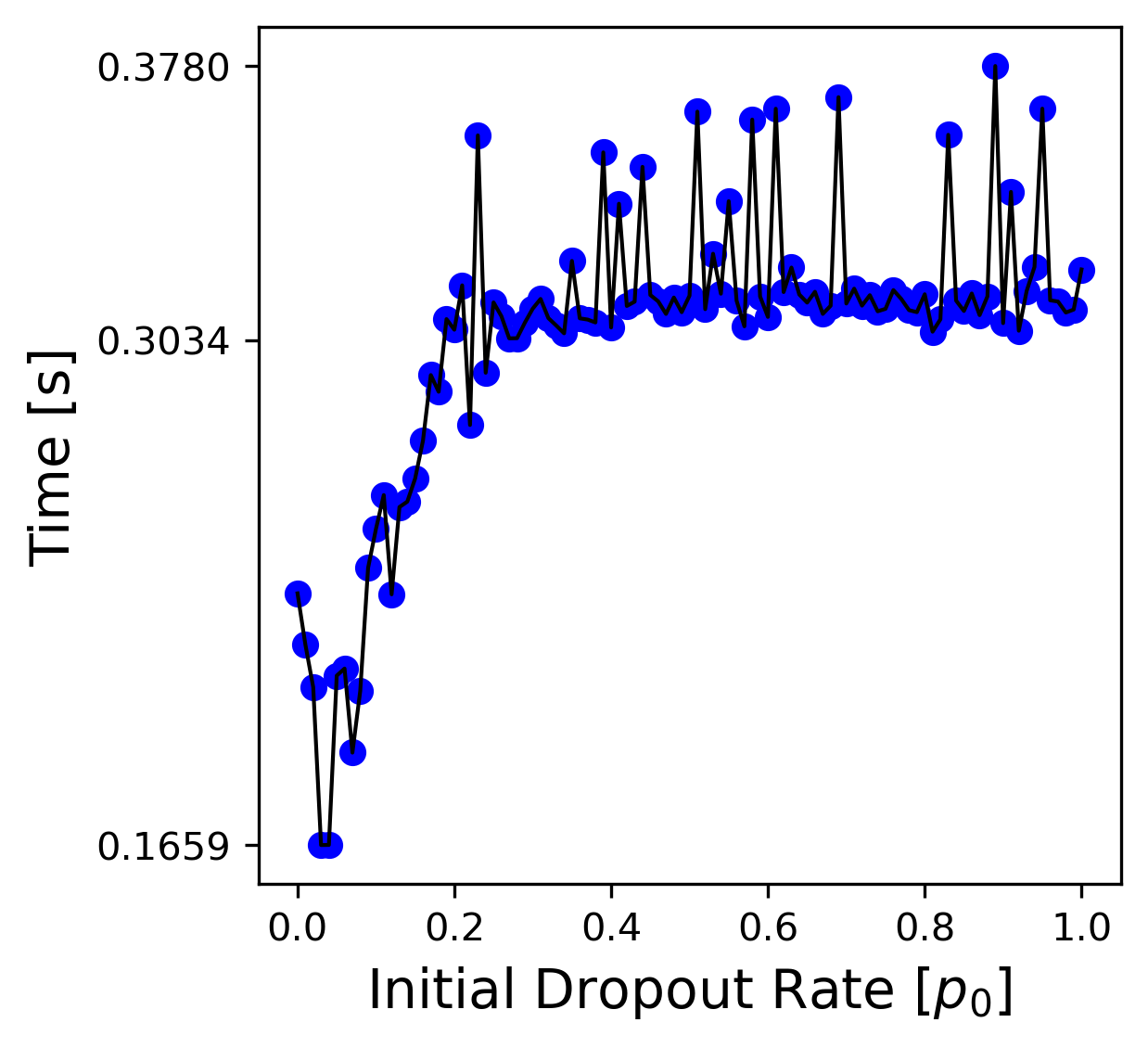}
            \caption{\scriptsize initial dropout rate}
            \label{fig:sub1_}
        \end{subfigure}
        \hspace{0.05\columnwidth}
        \begin{subfigure}{0.45\columnwidth}
            \includegraphics[width=\linewidth]{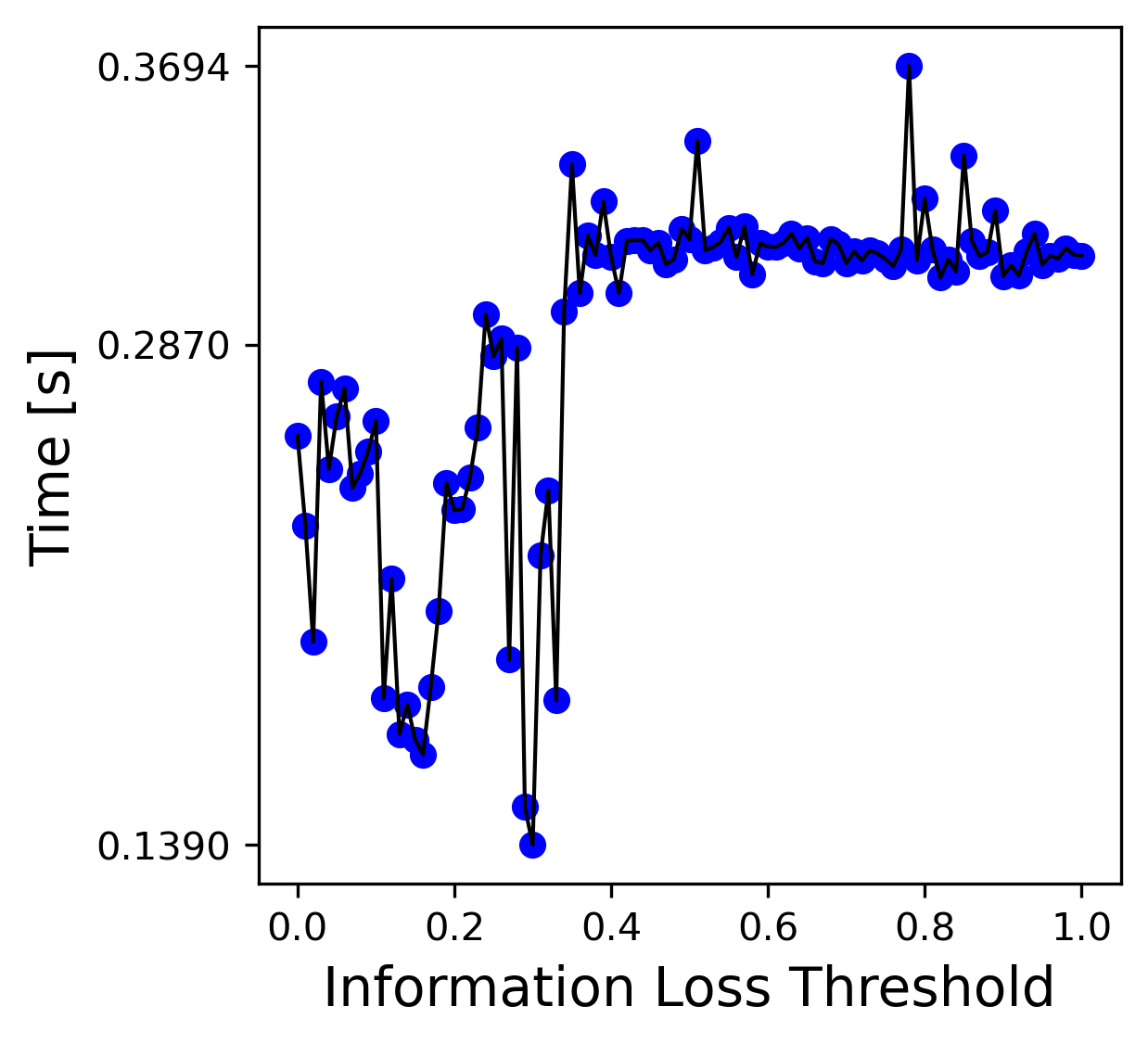}
            \caption{\scriptsize loss threshold}
            \label{fig:sub2}
        \end{subfigure}
    \end{minipage}
    
    \vspace{0.2em}
    
    \begin{minipage}{0.95\columnwidth}
        \centering
        \begin{subfigure}{0.45\columnwidth}
            \includegraphics[width=\linewidth]{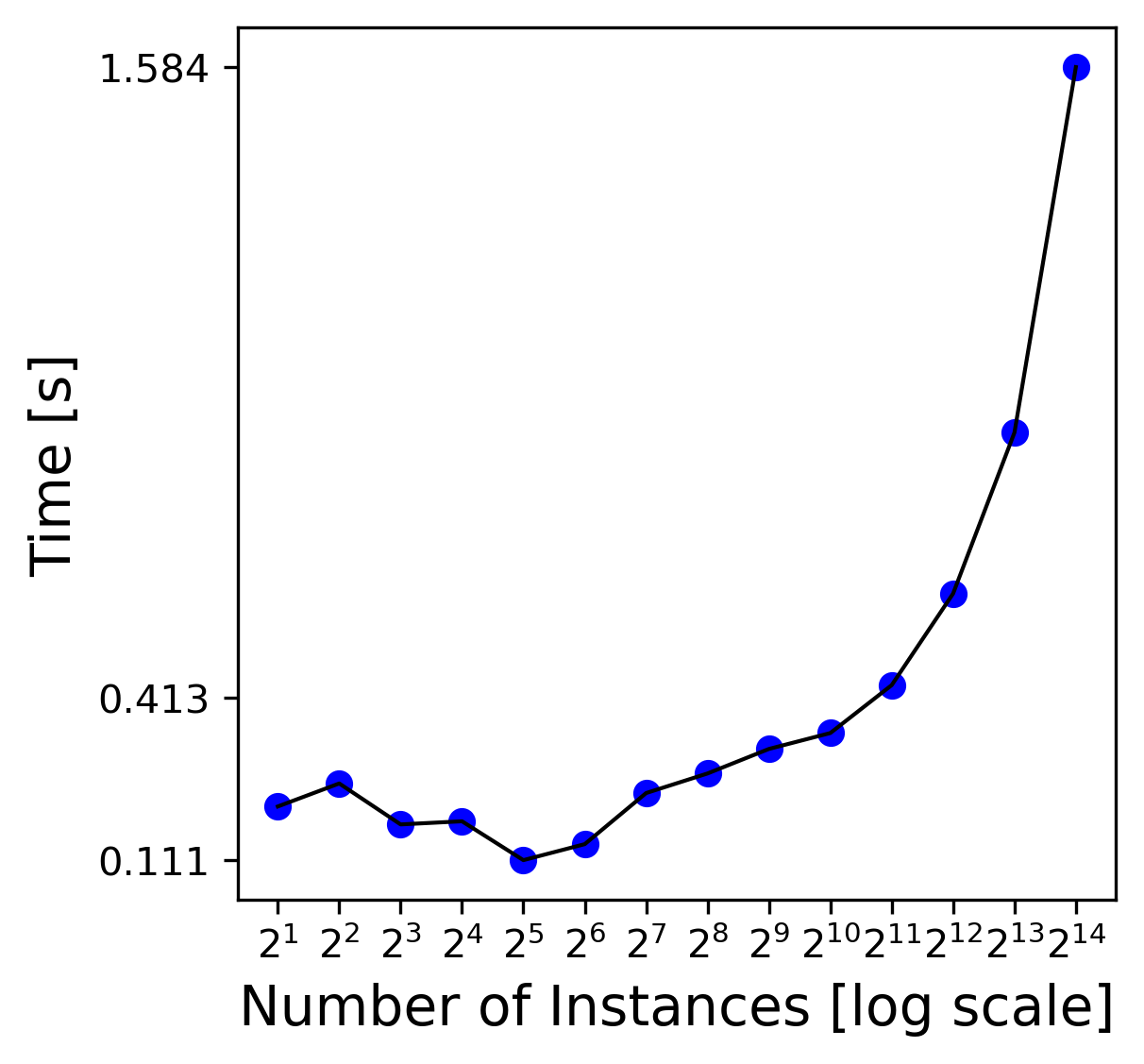}
            \caption{\scriptsize number of instances}
            \label{fig:sub3_}
        \end{subfigure}
        \hspace{0.05\columnwidth}
        \begin{subfigure}{0.45\columnwidth}
            \includegraphics[width=\linewidth]{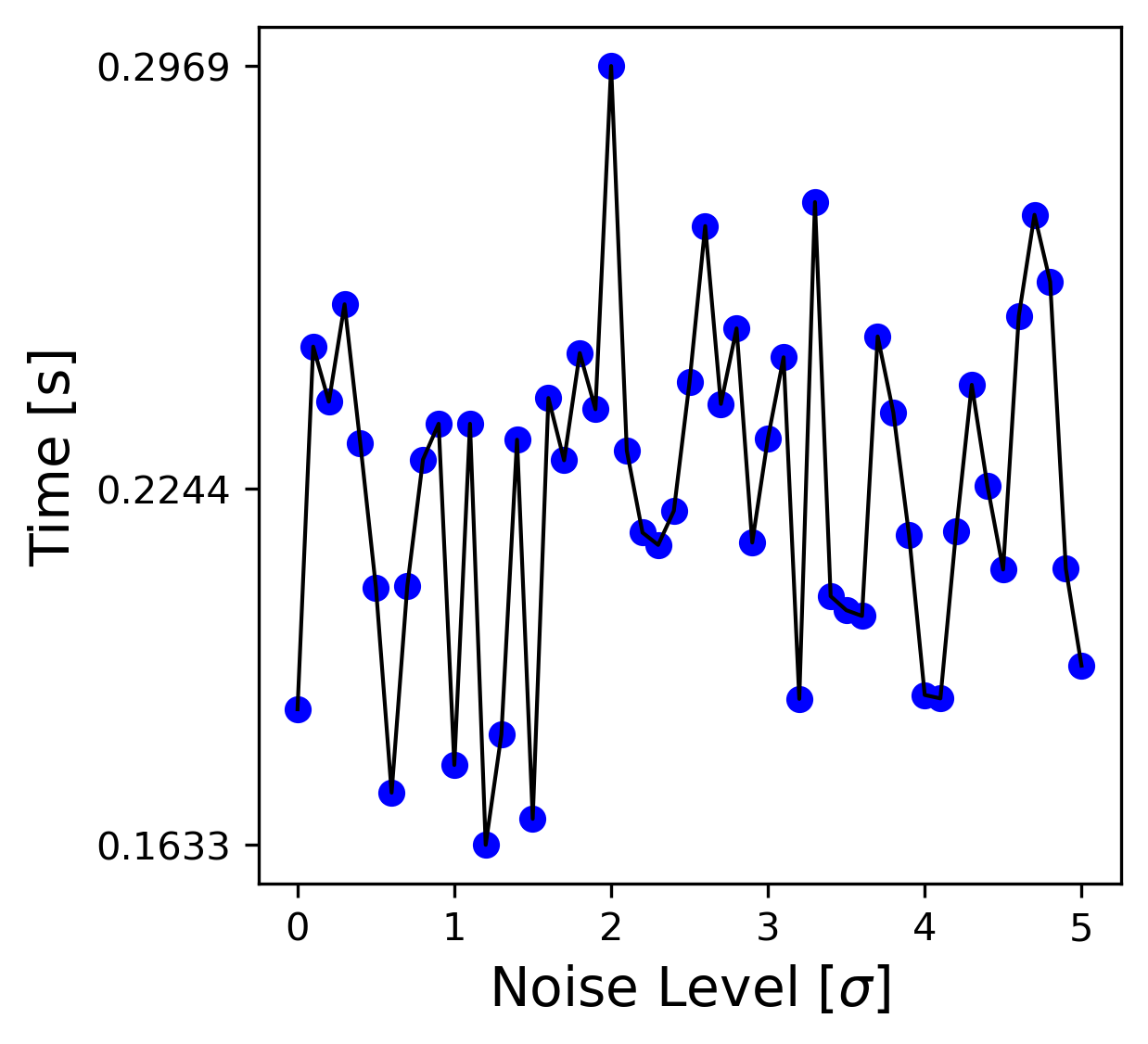}
            \caption{\scriptsize noise level}
            \label{fig:sub4}
        \end{subfigure}
    \end{minipage}
    
    \caption{\small \textbf{Rate-In execution time analysis for synthetic regression task}. Plots show optimization process duration (in seconds) as a function of: (a) initial dropout rate, (b) information loss threshold, (c) number of inference instances, and (d) noise level.}
    \label{fig:reg_comp_complexity}
\end{figure}

\paragraph{Synthetic Data:} Figures~\ref{fig:reg_comp_complexity} illustrate the time (in seconds) required to apply Rate-In to the synthetic regression task described in Section~\ref{sec:exp_setup}. We analyzed how Rate-In execution time varies as a function of four parameters: initial dropout rate ($p_0$), noise level ($\sigma$), loss threshold ($\epsilon$), and number of data instances. For each analysis, we varied one parameter while keeping the others fixed. The maximum number of Rate-In iterations ($N_{max}$) was set to 30, with $\delta = 0.001$ and a learning rate of 0.9 across all experiments.

Rate-In running time plateaus with increased initial dropout rates and information loss probabilities, while scaling linearly with training examples (note log-scale x-axis). Noise levels show minimal impact on execution time. The optimization process occasionally failed to find valid dropout rates in deeper layers, particularly in the second hidden layer. This occurred when the cumulative information loss from earlier layers made it mathematically impossible to achieve the target information loss threshold $\epsilon$, even with a dropout rate of zero. We defined convergence failure as the inability to reduce the objective function below the threshold $\epsilon$ after $N_{max}$ iterations, despite reaching the minimum possible dropout rate. Implementation of early stopping could mitigate computational costs.

\paragraph{Medical Data:} Tables ~\ref{tab:rate_in_duration_classification} and ~\ref{tab:rate_in_duration_segmentation} report the mean duration (with standard deviation) in seconds for Rate-In processing of individual instances in real-world classification and segmentation tasks for different values of information loss objectives. Initial dropout rate was set to be equal to the information loss objectives ($p_0 = \epsilon$), while all other parameters remained unchanged. The tables also include worst-case scenarios.

\begin{table}[ht]
\footnotesize
\centering
\caption{Mean (± std) and worst-case processing times (in seconds) of MI-based Rate-In for classification tasks.}
\label{tab:rate_in_duration_classification}
\setlength{\tabcolsep}{4.5pt}
\begin{tabular}{lllll}
\hline
\textbf{Dataset} & \textbf{Metric} & \textbf{$\varepsilon = 0.05$} & \textbf{$\varepsilon = 0.10$} & \textbf{$\varepsilon = 0.20$} \\
\hline
TissueMNIST & Mean (Std)  & 0.42 (0.07) & 0.43 (0.03) & 0.39 (0.12) \\
            & Worst-case  & 0.50        & 0.45        & 0.48        \\
PathMNIST   & Mean (Std)  & 0.65 (0.01) & 0.67 (0.02) & 0.56 (0.12) \\
            & Worst-case  & 0.66        & 0.68        & 0.65        \\
BloodMNIST  & Mean (Std)  & 0.68 (0.01) & 0.64 (0.01) & 0.48 (0.12) \\
            & Worst-case  & 0.70        & 0.65        & 0.57        \\
\hline
\end{tabular}
\end{table}

\begin{table}[ht]
\footnotesize
\centering
\caption{Mean (± std) and worst-case processing times (in seconds) of MI-based Rate-In for segmentation tasks.}
\label{tab:rate_in_duration_segmentation}
\setlength{\tabcolsep}{4.5pt}
\begin{tabular}{lllll}
\hline
\textbf{Dataset} & \textbf{Metric} & \textbf{$\varepsilon = 0.01$} & \textbf{$\varepsilon = 0.05$} & \textbf{$\varepsilon = 0.10$} \\
\hline
Prostate    & Mean (Std)  & 15.48 (5.48)  & 22.48 (7.48)  & 28.8 (9.55)   \\
            & Worst-case  & 27.43         & 40.06         & 48.72         \\
Liver       & Mean (Std)  & 10.08 (9.59)  & 28.62 (16.62) & 36.11 (18.74) \\
            & Worst-case  & 56.89         & 77.4          & 95.76         \\
\hline
\end{tabular}
\end{table}

\subsection{Structural Similarity Index (SSIM)}
The Rate-In algorithm enables domain-specific information loss measurements through user-defined metrics. While mutual information quantifies statistical dependencies affected by dropout noise, the SSIM Index ~\cite{wang2004image} can help evaluate spatial relationships in feature maps. This property makes SSIM relevant for CNNs, as it measures the preservation of local structures and patterns that these networks extract through their convolutional layers.

We used SSIM to measure information loss between pre- and post-dropout feature maps. This approach allows for simpler implementation as it only requires local comparisons within each layer, without needing access to the full model or input data. Based on earlier experiments showing non-negative SSIM values under dropout noise, the information loss at layer $l$ is defined as:
$\Delta I_l= 1 - \text{SSIM}(\mathbf{h}_{\text{drop}}^{(l)}{\text{in}}, \mathbf{h}_{\text{drop}}^{(l)}{\text{out}})$

\begin{figure}[ht]
    \centering
    \begin{subfigure}[b]{0.23\textwidth}
        \centering
        \includegraphics[width=\textwidth]{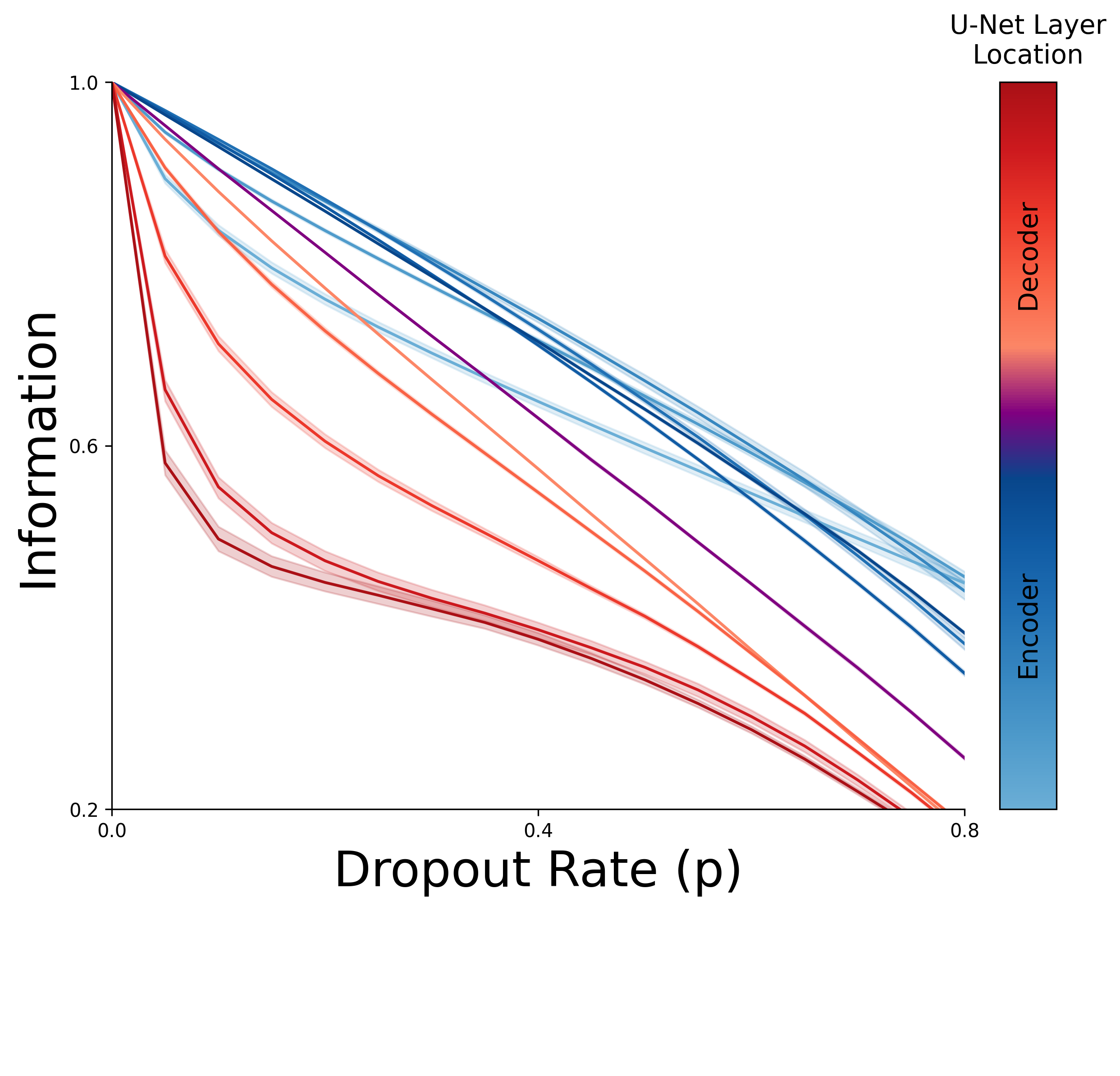}
        
        \caption{\scriptsize SSIM vs. Dropout Rate}
        \label{fig:ssim_vs_dropout}
    \end{subfigure}%
    \hspace{0.01\textwidth}%
    \begin{subfigure}[b]{0.23\textwidth}
        \centering
        \includegraphics[width=\textwidth]{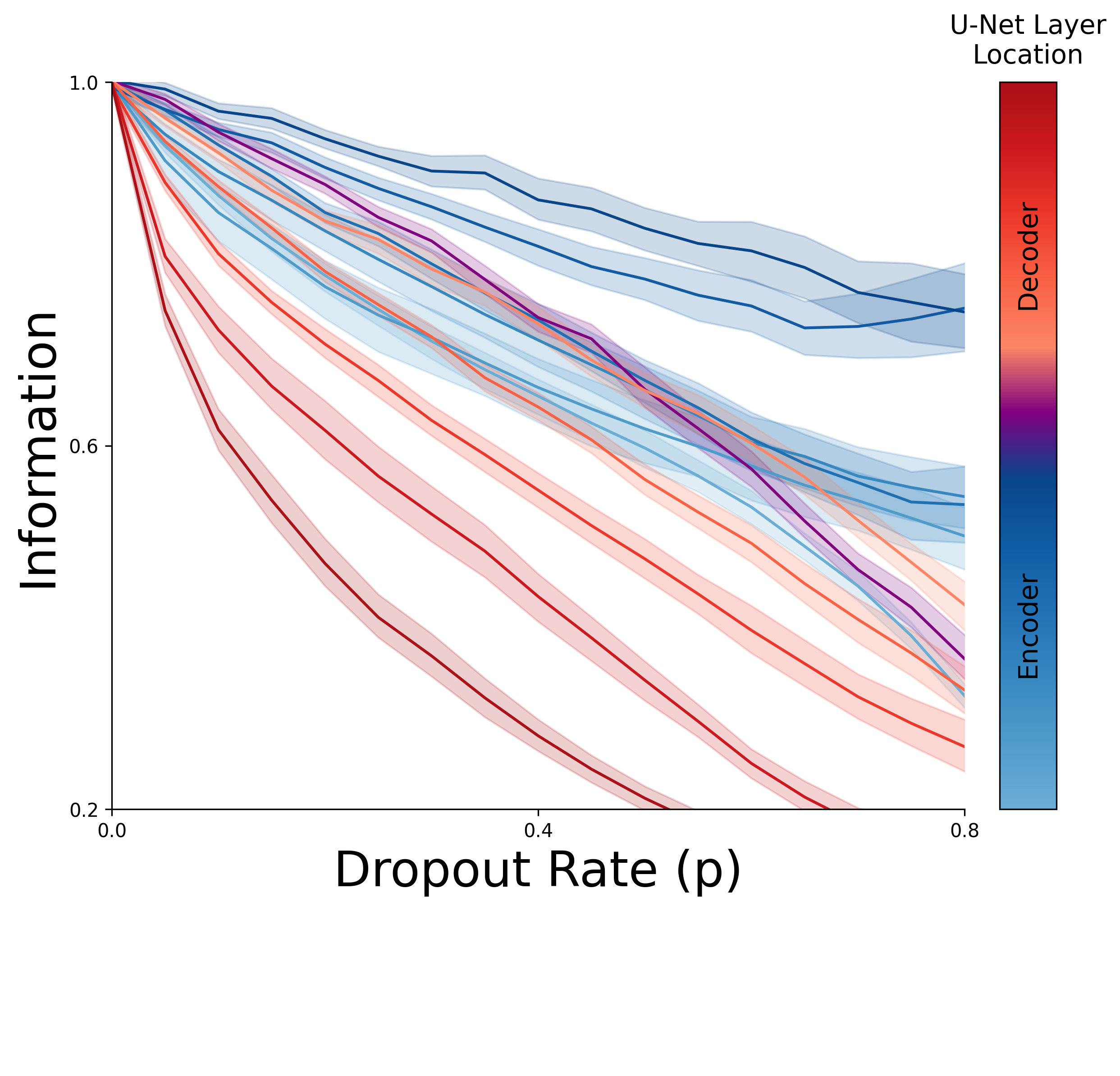}
        \caption{\scriptsize MI vs. Dropout Rate}
        \label{fig:mi_vs_dropout}
    \end{subfigure}

\caption{\small Dropout-induced information loss in the U-Net segmentation network as measured by SSIM (left) and MI (right) across different dropout rates and layers.}
    \label{fig:seg_sen_ssim}
\end{figure}

The performance comparison between SSIM-based and MI-based Rate-In across segmentation tasks is presented in Table ~\ref{tab:ssim_table}, with corresponding processing times for the SSIM approach shown in Table ~\ref{tab:rate_in_duration_segmentation_ssim}. To ensure consistent SSIM computation across different network layers, we normalized all feature maps to the [0,1] range. We set data\_range=1 in the torchmetrics SSIM implementation to match this normalized range, allowing meaningful comparison of structural similarities regardless of the original feature map magnitudes. Figure ~\ref{fig:seg_sen_ssim} shows SSIM (left) and MI (right) dropout-induced information loss across layers in the U-Net prostate segmentation network.

\begin{table}[ht]
\footnotesize
\centering
\caption{Mean (± std) and worst-case processing times (in seconds) of SSIM-based Rate-In for segmentation tasks.}
\label{tab:rate_in_duration_segmentation_ssim}
\setlength{\tabcolsep}{4.5pt}
\begin{tabular}{lllll}
\hline
\textbf{Dataset} & \textbf{Metric} & \textbf{$\varepsilon = 0.01$} & \textbf{$\varepsilon = 0.05$} & \textbf{$\varepsilon = 0.10$} \\
\hline
Prostate    & Mean (Std)  & 9.5 (1.53)     & 28.29 (4.59)   & 46.72 (7.23)  \\
            & Worst-case  & 13.86          & 44.07          & 71.9          \\
Liver       & Mean (Std)  & 11.39 (12.59)  & 13.08 (4.71)   & 17.13 (7.62)  \\
            & Worst-case  & 71.46          & 22.88          & 33.13         \\
\hline
\end{tabular}
\end{table}

\begin{table*}[t]
\caption{Medical image segmentation performance of Rate-In across anatomical regions in MRI and CT: comparing segmentation accuracy (DSC) and uncertainty calibration (ECE, BUC) using MI and SSIM-based information metrics at varying dropout rates.}

\label{tab:ssim_table}
\centering
\footnotesize
\setlength{\tabcolsep}{2.5pt}
\begin{tabular}{@{}lllcccccccccc@{}}
\toprule
\multirow{2}{*}{Zone} & \multirow{2}{*}{Modality} & \multirow{2}{*}{Method} & DSC & \multicolumn{3}{c}{DSC (Full Model, \% change)} & \multicolumn{3}{c}{ECE $\times 10^{-3}$} & \multicolumn{3}{c}{BUC} \\
\cmidrule(lr){5-7} \cmidrule(lr){8-10} \cmidrule(lr){11-13}
& & & Full Model & $p\text{,}\epsilon{\text{=}}0.01$ & $p\text{,}\epsilon{\text{=}}0.05$ & $p\text{,}\epsilon{\text{=}}0.10$ & $p\text{,}\epsilon{\text{=}}0.01$ & $p\text{,}\epsilon{\text{=}}0.05$ & $p\text{,}\epsilon{\text{=}}0.10$ & $p\text{,}\epsilon{\text{=}}0.01$ & $p\text{,}\epsilon{\text{=}}0.05$ & $p\text{,}\epsilon{\text{=}}0.10$ \\
\midrule
\multirow{2}{*}{Peripheral} & \multirow{2}{*}{MRI} & Rate-In (MI) & \multirow{2}{*}{0.682} & +0.29\% & +2.79\% & +2.35\% & 4.60 & \textbf{4.26} & \textbf{4.58} & \textbf{0.67} & \textbf{0.63} & \textbf{0.61} \\

& & Rate-In (SSIM) & & +3.05\% & +2.59\% & +1.82\% & \textbf{4.57} & 4.52 & 5.54 & 0.61 & 0.54 & 0.51 \\

\midrule
\multirow{2}{*}{Transitional} & \multirow{2}{*}{MRI} & Rate-In (MI) & \multirow{2}{*}{0.892} & -0.34\% & -0.78\% & -0.67\% & 3.97 & \textbf{4.03} & \textbf{4.86} & \textbf{0.87} & \textbf{0.84} & \textbf{0.80} \\

& & Rate-In (SSIM) & & -0.72\% & -0.76\% & -0.87\% & \textbf{3.95} & 4.78 & 6.72 & 0.81 & 0.72 & 0.67 \\

\midrule
\multirow{2}{*}{Liver} & \multirow{2}{*}{CT} & Rate-In (MI) & \multirow{2}{*}{0.955} & +1.05\% & +1.05\% & +0.94\% & \textbf{6.73} & 5.54 & 4.65 & 0.93 & 0.93 & 0.92 \\

& & Rate-In (SSIM) & & +1.82\% & +1.16\% & +1.18\% & 6.82 & \textbf{5.49} & \textbf{4.47} & 0.93 & 0.93 & 0.92 \\

\midrule
\multirow{2}{*}{Tumor} & \multirow{2}{*}{CT} & Rate-In (MI) & \multirow{2}{*}{0.579} & +1.21\% & -0.52\% & -2.59\% & \textbf{1.78} & 1.78 & 1.88 & 0.61 & 0.53 & 0.48 \\

& & Rate-In (SSIM) & & +1.32\% & +1.29\% & +1.29\% & 1.81 & \textbf{1.76} & \textbf{1.74} & \textbf{0.64} & \textbf{0.60} & \textbf{0.55}\\

\bottomrule
\end{tabular}
\end{table*}

\section{Appendix C: Further Discussion}
\label{suppl:appendix_C}

\paragraph{Dropout, beyond just graph manipulation.} Rather than viewing dropout as merely a tool for graph manipulation, Rate-In reinterprets it as a method for controlled noise injection. This perspective shift allows us to examine how dropout affects network representations in task-specific contexts. Rate-In effectively acts as a 'noise translator', converting random dropout noise into meaningful, task-specific variations in feature maps. Similar to how test-time augmentation adds meaningful noise to inputs, Rate-In dynamically manages noise levels in feature maps to achieve more accurate uncertainty estimates.

\paragraph{Rate-In, enhancing dropout post-training.} Nearly a decade after its introduction, classical Bernoulli dropout is still among the most popular regularization techniques in deep learning, outlasting many newer alternatives. Rate-In builds upon this foundation by introducing dynamic rate adjustment without modifying the underlying dropout mechanism, making it easy for users to adopt Rate-In without altering their existing workflows.

\paragraph{Similar dropout rates, different effects: layer position matters.} Our experiments revealed that dropout's impact varies significantly across network layers, even when feature map dimensions are similar. In semantic segmentation tasks, we found that decoder layers showed higher sensitivity to dropout rate changes compared to encoder layers. This variation may be attributed to MRI image characteristics - uniform dark backgrounds in early layers mitigate dropout's impact, while intensity variations in later layers amplify it. In addition, decoder layers have fewer opportunities to compensate for dropout noise, emphasizing the importance of layer-specific rate adjustment.

\paragraph{Sequential dropout adjustments: preserving inter-layer rate dependency} The Rate-In algorithm operates sequentially during the forward pass of a pre-trained network. At each dropout layer, a feedback loop adjusts the dropout rate to align with the information loss objectives specific to that layer. After adjustment, the modified information propagates through the network to the next dropout layer, where the process repeats. This approach enables control over information loss across the entire network, establishing a dropout rate policy that maintains inter-layer dependency.

\paragraph{Additional complexity in inference, but can be handled offline.} Rate-In introduces additional computation at inference. While this can be challenging for some real-time applications, a few potential solutions can be considered. Rate-In can be applied offline to a dataset, such as the training or validation set, to establish population-level dropout rates that approximate individual case needs, assuming data distribution alignment. These rates can then serve as starting points for further tuning during inference or be sampled from the population rate distribution for each layer.

\subsection{Limitations}
\label{limitations}

\paragraph{MI might not fully capture vision task nuances.} In our experiments, we used mutual information (MI) to quantify functional information loss from dropout. While MI is a common choice, it may not capture all aspects of information critical in computer vision tasks. Supplementing MI with metrics like the Structural Similarity Index (SSIM) could provide a more comprehensive assessment, covering different dimensions of information quality and enabling more precise dropout rate adjustments to address various types of potential information loss.

\paragraph{Loss objectives may need empirical validation.} Rate-In’s feedback loop relies on an empirically defined information loss objective. Our experiments suggest that keeping information loss below 10\% preserves functional information while retaining dropout’s benefits. Currently, each dropout layer applies a local loss function, resulting in distinct dropout rates per layer. The current implementation uses independent loss functions for each dropout layer, leading to layer-specific dropout rates. One potential modification would be to use a network-wide information loss constraint - a 'dropout budget' shared across layers.

\paragraph{Dropout does not cover all types of noise.} In convolutional layers, dropout adds noise only within small, localized regions defined by the convolution kernel. As a result, it may not capture broader noise patterns such as frequency artifacts, intensity shifts, or large-scale texture variations that span entire images. This limitation restricts dropout’s effectiveness in addressing global noise types that are more uniformly spread across feature maps.

\end{document}